\documentclass[10pt,twocolumn,letterpaper]{article}

\usepackage{wacv}
\usepackage{times}
\usepackage{epsfig}
\usepackage{subfig}
\usepackage{graphicx}
\usepackage{amsmath}
\usepackage{amssymb}
\usepackage{bbm}
\usepackage{bbold}
\usepackage{todo}
\usepackage[ruled,noend]{algorithm2e}
\usepackage{float}
\usepackage{enumitem}
\long\def\comment#1{}
\newtheorem{definition}{Definition}
\usepackage{dirtytalk}
\input{tbspace}

\def\wacvPaperID{1263} 
\wacvfinalcopy 

\ifwacvfinal
\def\assignedStartPage{1} 
\fi
\ifwacvfinal
\else
\fi

\ifwacvfinal
\else
\fi

\begin{document}

\title{Automatic Open-World Reliability Assessment}

\author{Mohsen Jafarzadeh \;\;\;\;\; Touqeer Ahmad \;\;\;\;\; Akshay Raj Dhamija\\
Chunchun Li \;\;\;\;\;\;\;\;\;\;\;\;\; Steve Cruz \;\;\;\;\;\;\;\;\;\;\; Terrance E. Boult
 \thanks{This research was  sponsored  by the Defense Advanced Research Projects Agency (DARPA)  under HR001120C0055. The views contained in this document are those of the authors and should not be interpreted as representing the official policies, either expressed or implied, of the DARPA or the U.S. Government.}
\\
University of Colorado, Colorado Springs\\
Colorado Springs, Colorado 80918, USA\\
{\tt\small \{mjafarzadeh, tahmad, adhamija, cli, scruz, tboult\}@vast.uccs.edu}
}

\maketitle
\tightmath
\begin{abstract}

Image classification in the open-world must handle out-of-distribution (OOD) images. Systems should ideally reject OOD images, or they will map atop of known classes and reduce reliability. Using open-set classifiers that can reject OOD inputs can help. However, optimal accuracy of open-set classifiers depend on the frequency of OOD data.
Thus, for either standard or open-set classifiers, it is important to be able to determine when the world changes and increasing OOD inputs will result in reduced system reliability. However, during operations, we cannot directly assess accuracy as there are no labels. Thus, the reliability assessment of these classifiers must be done by human operators, made more complex because networks are not 100\% accurate, so some failures are to be expected.  To automate this process, herein, we formalize the open-world recognition reliability problem and propose multiple automatic reliability assessment policies to address this new problem using only the distribution of reported scores/probability data. The distributional algorithms can be applied to both classic classifiers with SoftMax as well as the open-world Extreme Value Machine (EVM) to provide automated reliability assessment. We show that all of the new algorithms significantly outperform detection using the mean of SoftMax.

\end{abstract}

\begin{figure}[t!]
\centering
\subfloat[SoftMax for 6 batches]{\includegraphics[width=.45\linewidth]{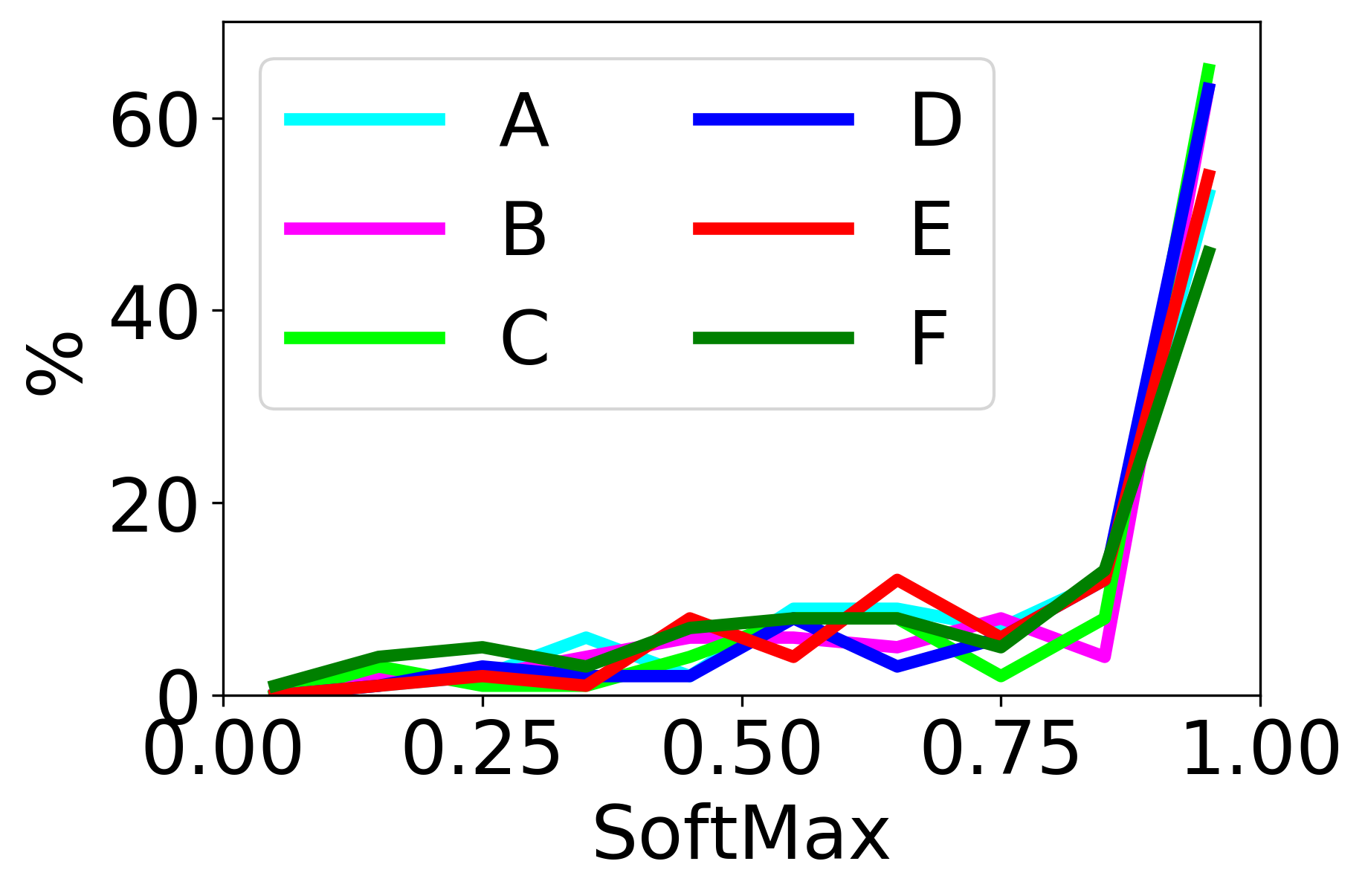}}
\subfloat[EVM for 6 batches]{\includegraphics[width=.45\linewidth]{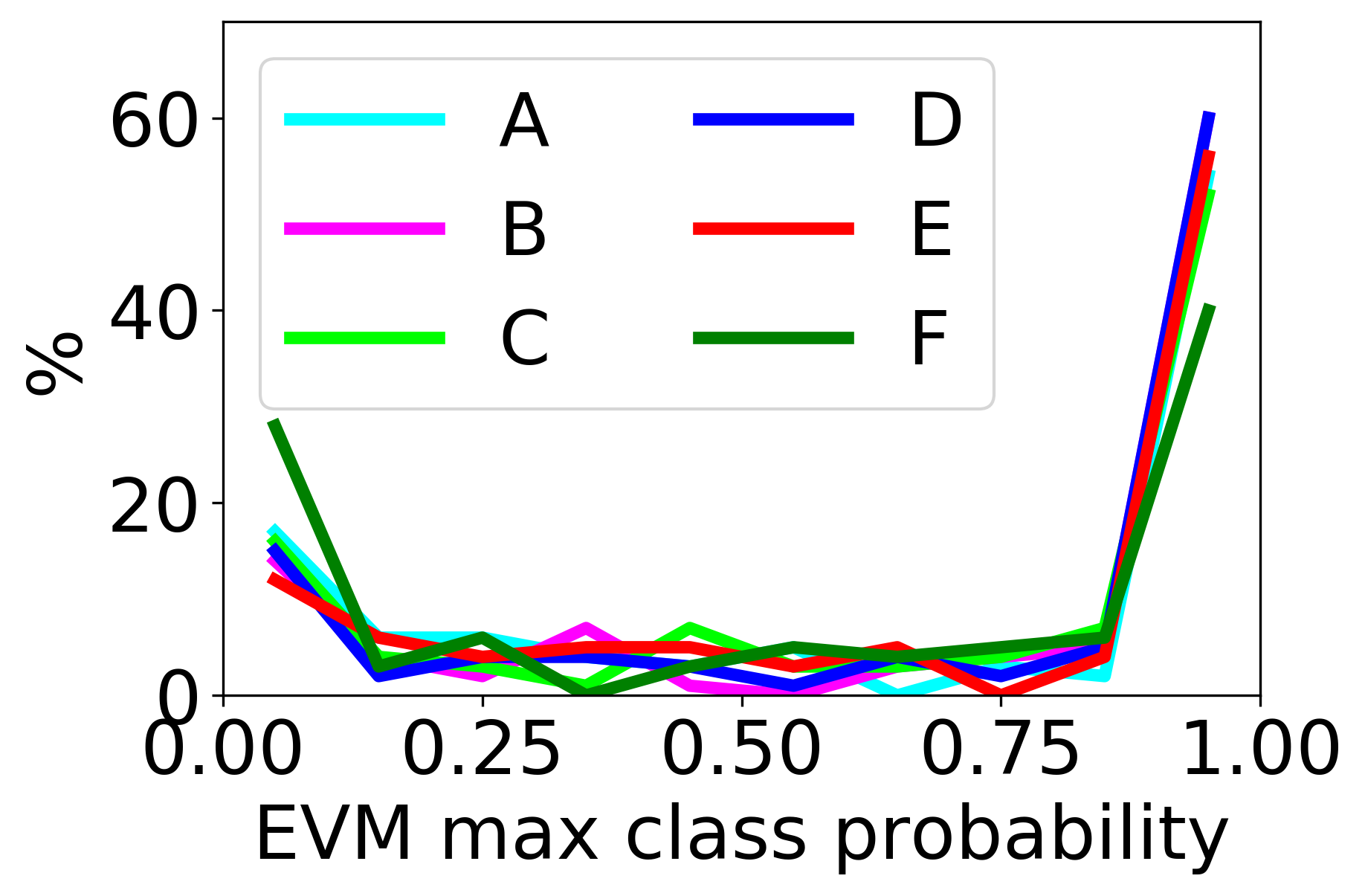}}\\
\subfloat[Scores over different data partitions]{\includegraphics[width=.8\linewidth]{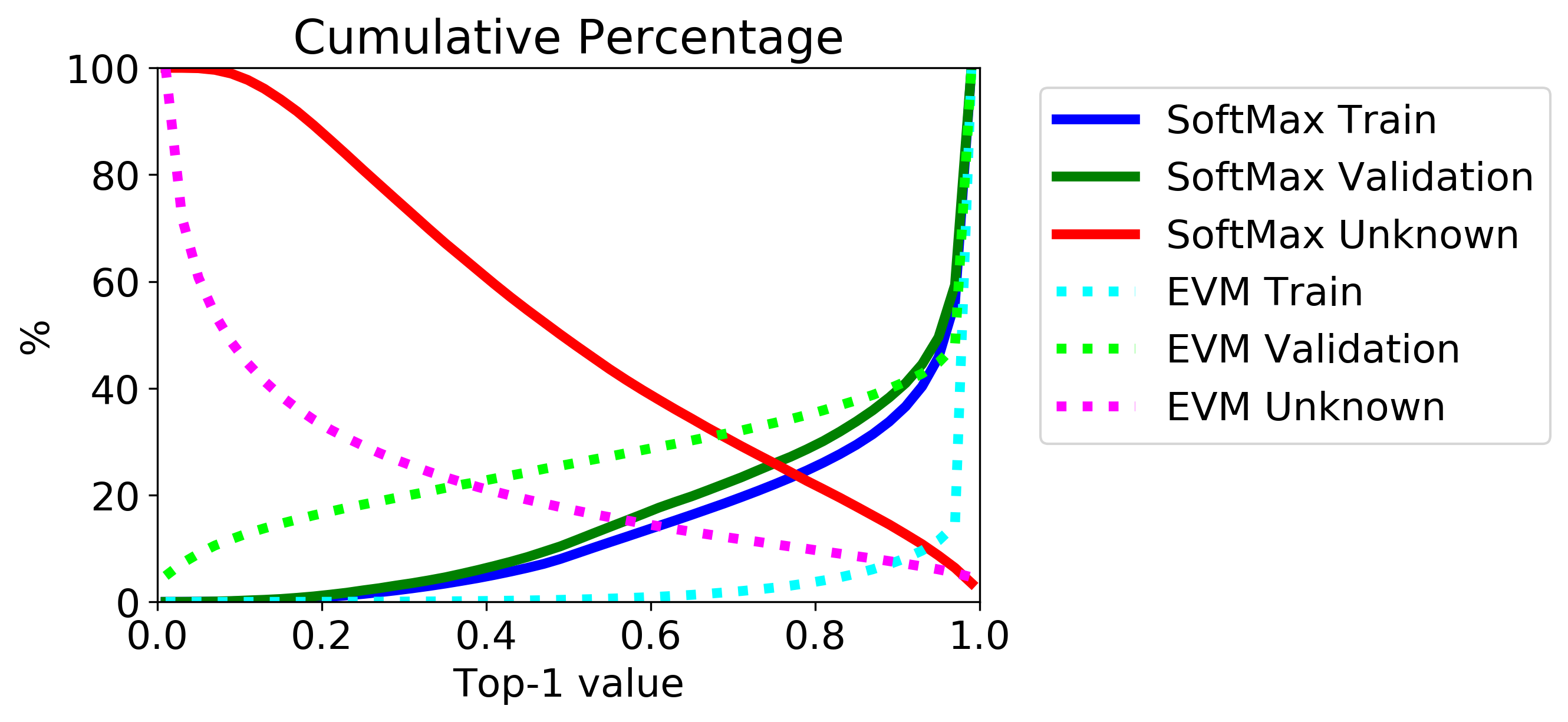}}\\
\caption{Percentage of images that yield the given score/probability for either SoftMax or EVM. In (a) and (b), we show the percentage for six different batches of 100 images. Reliability is about to determine when the system's accuracy will be degraded by OOD samples. Can you determine which of the six batches (A--F) are "normal" and which have increased OOD data? As a hint, three are normal; the others have 3\%, 5\%, and 20\% OOD data. See the text for answers. Open-set classifiers deal with OOD inputs rejecting inputs below a  threshold on score/probability.  Plot (c) shows the cumulative percentage of images above threshold for knowns and below threshold for unknowns, i.e., the percentage of data correctly handled at that threshold. Assessing a change in reliability,  i.e., detecting a shift in the distribution of knowns and unknowns, is a non-trivial problem because of the overlap of scores on knowns and OOD images. This paper formalizes and explores solutions to the open-world reliability problem.}
\label{fig_distribution}
\end{figure}

\section{Introduction }

Many online businesses, especially shopping websites and social media, created a platform that users can upload an image, and it classifies the images into predefined classes, potentially open-set classifiers, which in addition to known classes in the training data set, have an out-of-distribution (OOD) class, also known as novel, garbage, unknown, reject, or background class, that the system otherwise ignores. The OOD class can be considered as a "background" separate class or as the origin of feature space \cite{dhamija2018reducing}. 
Mapping too many images to reject class or misclassifying them reduces users' satisfaction. But since there are no labels during operation, there is no obvious way to detect that the input distribution has changed. 
Therefore, online businesses use human laborers to monitor image classification quality and to fine-tune the system when the quality drops.
Monitoring using human labor is expensive.
In other applications, e.g., robotic or vehicular navigation, reduced reliability from increasing OOD inputs must be handled automatically as the processing cycle is too fast for human oversight. Fig.~\ref{fig_distribution} shows score distributions and why thresholding can be problematic, and why detecting changes in the mix of known and OOD is difficult. Did you correctly guess that A, B, and C are the normal batches while D, E, and F have 3\%, 5\%, and 20\% OOD data, respectively?

As we move toward the path of operational reliability, we start by adding robustness to close-set classifiers. We can further improve reliability by transitioning to open-set classifiers or even transition to open-world classifiers that incrementally add the OOD samples \cite{boult2019learning}. 
While there has been a modest amount of research looking at various definitions of classifier reliability determination (see Section~\ref{s:related}) and defining open-set and open-world recognition algorithms, no prior work has addressed the problem of  reliability when there are changes in the frequency of OOD inputs.

In this work, we take operational reliability one step further by defining and presenting solutions to the problem of automatically assessing classifier reliability in an open-set/open-world setting. Such assessments are necessary to determine when it may be time to fine-tune and incrementally learn new classes; however, the choice of thresholds and processes for incremental learning is domain-specific and beyond the scope of this paper.

Extreme value theory (EVT) is a branch of statistics that studies the behavior of extreme events, the tails of probability distributions \cite{coles2001introduction, beirlant2006statistics, castillo2012extreme}. EVT estimates the probability of events that are more extreme than any of the already observed ones. EVT is an extrapolation from observed samples to unobserved samples. The extreme value machine (EVM) \cite{rudd2017extreme, henrydoss2017incremental} uses EVT to compute the radial probability of inclusion of a point with respect to the class of another. Therefore, EVM has been successfully used for instance-level OOD detection.

While the EVM is an inherently open-set classifier, it is less widely used, and hence it is also natural to ask if we can detect when there is a shift in the distribution for a standard closed-set classifier and hence the need for retraining. In this paper, we show distribution-based algorithms, including Gaussian-model with  Kullback–Leibler (KL) divergence applied to SoftMax scores or EVM scores, are effective in detection. We also develop a novel detection algorithm for EVM, which makes Bernoulli distributional assumptions and find it works equally as well. Thus, the proposed reliability algorithms can be used for both traditional "closed-set" and more recent "open-set" classifiers for open-world reliability assessment.

\vspace*{-2ex}
\paragraph{Our Contributions}
\begin{itemize}[nosep]
\item Formalizing open-world reliability assessment.
\item Proposing three new policies to automate open-world reliability assessment of closed-set (SoftMax) and open-set image classifiers (EVM).
\item Developing a test-bed to evaluate the performance of reliability assessment policies.
\item Comparing proposed policies with simple approaches, showing a significant improvement over the base algorithm of tracking the mean of SoftMax scores. 
\end{itemize}

\section{Related work}
\label{s:related}
The robustness of image classifiers has been studied for many years, with a known trade-off between the accuracy and robustness of image classifiers \cite{su2018robustness}. The effect of data transformation in robustness is demonstrated in \cite{bhagoji2018enhancing}. Stability training is another way of increasing robustness \cite{zheng2016improving}. Although, by improving robustness, an image classifier can increase reliability to variation of known classes; in practice, robustness degrades when the ratio of OOD images to other images increases.

When presented with an out-of-distribution input, most image classifiers map OOD to a known class. Recently introduced open-set classifiers \cite{bendale2016openmax,rudd2017extreme,yoshihashi2019classification,oza2019c2ae,bhattacharjee2020multi,geng2020collective,leng2019survey}, have developed improved approaches to reasonably detect if an input is from a novel class (a.k.a unknown or out-of-distribution) and map it to a reject or novel class label. 
A family of quasi-linear polyhedral conic discriminant is proposed in \cite{cevikalp2019polyhedral} to overcome label imbalance. An iterative learning framework for training image classifiers with open-set noisy labels is proposed in \cite{wang2018iterative}. Networks can be trained for joint classification and reconstruction \cite{yoshihashi2019classification}. Class conditioned auto-encoders proposed may improve the accuracy of known while rejecting unknowns \cite{oza2019c2ae}. Policies for a guided exploration of the unknown in the open-world are proposed in \cite{liu2019large}. The robustness of open-world learning has been studied in \cite{sehwag2019analyzing}. Some techniques, such as \cite{rudd2017extreme}, allow incremental updating to support open-world learning of these new unknowns but provide no guidance on when the system's reliability has been reduced to the point of needing incremental updates.
The quality of these open-set image classifiers is directly related to the ratio of known to unknown images. By decreasing this ratio, the open-set image classifiers should be fine-tuned to have better reliabilities on the data. Thus, businesses should invest in the reliability assessment policy.

There have been various studies of the reliability of classifiers \cite{liu2005one, zou2009effects,dantcheva2011search} that looked at close-set classifiers' accuracy or error.
These do not apply in the open-world setting.
Still others considered the largest few scores (extreme values) in the close-set setting \cite{zou2010discriminability,sun2013iris,hazelhoff2013robust} weak approximation of using extreme value for open-set classification as in \cite{bendale2016openmax}.
In \cite{ma2015new}, reliability is defined as the difference between the classification confidence values.
Matiz and Barner defined reliability as confidence values of predictions \cite{matiz2019inductive}.
In a published paper \cite{latifi2020polygraphmr}, authors showed that confidence is not equal to reliability because image classifiers may put images in the wrong class with high confidence.
Any reliability measure on a single input can be viewed as a type of classifier with rejection, which is a subset of open-set recognition algorithms.

Looking at sets of data, either in batches or moving windows, it is necessary to detect the change in the distribution associated with increasing OOD samples.
In \cite{romani2005reliability}, authors defined reliability as an inverse to its standard deviation, i.e., $R = \sigma^{-1}$.
Zhang defined reliability as the sum of rejection rate and accuracy \cite{zhang2012reliable}.
This definition is not acceptable because if the image classifier puts all images in the reject class, the rejection rate is $1.0$, and consequently, the reliability is $1.0$.
In \cite{latifi2020polygraphmr}, they used a false positive rate of points where no desirable correct predictions are lost.
Their definition of reliability also fails for the same reason, if an image classifier says every image is negative (reject class), then it has a reliability of 1.0.

Considering the data as a sequence in time, one can view the reliability assessment problem as a specialized type of distributional change detection. In the change detection literature, there is a wide range of algorithms, and an important algorithm is KL divergence  \cite{dasu2009change}. 

Thus, while there has been a modest amount of research looking at various definitions of classifier reliability determination, no prior work addresses it from the point of view of open-set classifiers when there are changes in the frequency of OOD inputs.

\section{Problem Formalization \& Evaluation}

British standards institution defined reliability as “the characteristic of an item expressed by the probability that it will perform a required function under stated conditions for a stated period of time” \cite{dummer1997elementary}.
In this paper, we assume a well-trained network, such that per-class errors are relatively consistent across classes, i.e., the network is trained such that their errors are not a function of the classes.
With that assumption and the standard definition, networks should be reasonably reliable to changes in sampling frequency between classes in their training distributions, so distributional shifts between such classes are not important.
However, since OOD data is not something we can use in training, we cannot assume the uniformity of errors on OOD data and networks. Classifiers become unreliable if the frequency of samples from unseen classes change, and they increase their frequency of rejection.
Thus, we seek to formalize reliability via changes in the distribution with respect to OOD samples.

\begin{definition}{Open-world Reliability Assessment}

Let us define ${\cal T}$ for the mixture of known and unknown classes seen in training, as well as $\cal U$, classes unseen in training.
Let $x_1 \ldots x_N,$ be samples drawn from distribution ${{\cal D}}_1$, where $x\in{{\cal D}}_1 \rightarrow x \in {\cal T}$, i.e. ${\cal D}_1$ is either consistent with training or some operational validation set and hence is by definition reliable. 
Let $x_{N+1} \ldots x_m,$ be samples drawn from another distribution ${{\cal D}}_2$.
Let $P_{\cal U}(x;{{\cal D}})$ be the probability, given distribution $D$, that $x \in \cal U$, i.e., the probability $x$ is a class unseen in training.
Finally, let ${\cal M}$ be a dissimilarity measure between two probability distributions. 

{\underline {Open-world Reliability Assessment}} is determining if there is a change in the distribution of samples from the unknown or OOD classes, i.e. 
\begin{equation}
{\cal M}\bigl({ P_{\cal U}(x;{{\cal D}_2}) ; P_{\cal U}(x;{{\cal D}_1})}\bigr) > \delta
\label{eq:ra}
\end{equation}
for some user-defined $\delta$. When the above equation holds, we say that ${\cal D}_2$ has a change in reliability.
\end{definition}

The choice of $\cal M$ and $\delta$ may be domain-dependent as the cost of errors may be influenced by the size and frequency of the distributional shifts. In this paper, we explore multiple models. As defined, this does not say if reliability is increasing or decreasing, just that it changed. However, with some measures, one can determine the direction of the change. 

From an experimental point of view, we can introduce different distributions ${\cal D}_2$ and measure detection rates
based on the ground truth label of when it was introduced. 

Given a sequence of samples, detection of the distributional change from reliable to unreliable is often an important
operational criterion. We define reliability localization error as the absolute error in the reported location of
reliability change, i.e., if the detection is reported at time $n_r$ and ground truth is $n_g$, for a given trial absolute error is 
\begin{equation}
e_l = | n_g - n_r|
\label{eq:mae}
\end{equation}
and can consider the mean absolute error over many trials. We define {\em On-Time detection} as reporting the change
with zero absolute error for a given trial, and early and late detection have a non-zero error with the appropriate sign.

To implement such an assessment, we need some way to estimate the probability of something being out of distribution. One simple and common approach is using classifier confidence, i.e., SoftMax, as in \cite{ma2015new}.  Better OOD detection approaches are based on the more formal open-set classifiers that have been developed, and for this work, we consider the model from the extreme value machine (EVM) \cite{rudd2017extreme,henrydoss2017incremental}.  While EVM may be better for individual OOD detection, we show the proposed reliability assessment approaches works equally well for both and expect it will work well with building on top of any good per-instance estimate of the probability of input being out-of-distribution.

\subsubsection*{Evaluation Protocol}
Similar to testing open-set classifiers, testing of open-world reliability assessment needs a mixture of in-distribution classes that was used during training as well as out-of-distribution data. For measuring open-set evaluation, we recommend a large number of classes to ensure good features; a small number of classes produce more degenerate features from overfitting. For this evaluation, we used all 1000 classes of ImageNet 2012 for training and validation as known classes. Then, to create a test set, we used combinations of 1000 classes of ImageNet 2012 validation set as known classes and 166 non-overlapping classes of ImageNet 2010 training set as unknown classes. We note that while some earlier work, such as \cite{bendale2016openmax,rudd2017extreme}, used 2012/2010 splits, they claimed 2010 has 360 non-overlapping class. While it may be true by class name, our analysis showed many super-class/subclass relations or renamed classes, e.g., ax and hatchet. To determine non-overlapping classes, we first processed 2010 classes with a 2012 classifier to determine which classes had more than 10\% of its test images classified within the same 2012 class. We then had a human check if the classes were the same. The final list has 166 classes and is in our GitHub  repo at https://github.com/ROBOTICSENGINEER/Automatic-Open-World-Reliability-Assessment.

Testing uses different percentages of data from the 2012 validation test data as knowns and from the 2010 classes as the unknowns or out-of-distribution data.
To evaluate the policies, we created 24 configuration tests. Each configuration has 10,000 tests. Each test consists of 4000 images. The first 2000 images are mostly knowns, 20 of them unknowns (1\%). The ratio of unknown images increased in the second 2000 images, ranging from 2\% to 25\% by 1\% increments. We used a batch size of 100.

\section{Background on EVM}

Because we use the EVM as a core of multiple new algorithms, we briefly review its properties, so this paper is self-contained. The EVM is a distance-based kernel-free non-linear classifier that uses Weibull families distribution to compute the radial probability of inclusion of a point with respect to the class of another. Weibull families distribution is defined as 
\begin{equation}
\label{eq_Weibull}
\textup{W} (x; \mu , \sigma , \xi) = 
\begin{cases}
e^{- (1 + \xi(\frac{x - \mu}{\sigma}))^{\xi}} & , x< \mu - \frac{\sigma}{\xi} \\ 
1 & , x \geq \mu - \frac{\sigma}{\xi} 
\end{cases}
\end{equation}
where $\mu \in \mathbb{R}$, $\sigma \in \mathbb{R}^+ $, and $\xi \in \mathbb{R}^-$ are locations, scale, and shape parameters \cite{coles2001introduction, beirlant2006statistics, castillo2012extreme}. 
EVM provides a compact probabilistic representation of each class’s decision boundary, characterized in terms of its extreme vectors. Each extreme vector has a family of Weibull distribution.s Probability of a point to each class is defined as the maximum probability of point belonging to each extreme vector of the class. 
\begin{align}
\label{eq_EVM_p}
\hat{P} (C_l|x) &= \max_{k} \textup{W}_{l,k} (x; \mu_{l,k} , \sigma_{l,k} , \xi_{l,k}) 
\end{align}
where $\textup{W}_{l,k} (x)$ is Weibull probability of $x$ corresponding to $k$ extreme vector in class $l$. If we show unknown label with 0, the predicted class label $\hat{y} \in \mathbb{W}$ can be computed as following.
\begin{equation}
\label{eq_m}
m(x) = \max_{l} \hat{P} (C_l|x)
\end{equation}
\begin{equation}
\label{eq_label}
z(x) = \textup{arg}\max_{l} \hat{P} (C_l|x)
\end{equation}
\begin{equation}
\label{eq_Heaviside}
q (x; \tau) = \textup{Heaviside}(m(x) - \tau)
\end{equation}
\begin{equation}
\label{eq_evm_y}
\hat{y} (x; \tau) = q (x; \tau) \; z(x)
\end{equation}
where $\tau \in \mathbb{R}^+$ is a threshold that can be optimized on some validation set. If the threshold is too large, EVM tends to put known images into the reject class, i.e., high false rejection rate. If the selected threshold is too small, EVM classifies unknown images to the known classes, i.e., high false classification rate. Therefore, this threshold acts as a slider in the trade-off between false rejection rate and false classification rate.

\section{ Algorithms for Open-world Reliability Assessment}

We process batches, or windows of data, and attempt to detect the first batch where the distribution has shifted.
Given the definition, the natural family of algorithms is to estimate the probability of input being unknown and then estimate properties of the distributions and compare them.

\paragraph{Mean of SoftMax}
Thresholding the SoftMax score of neural networks is well-known and the simplest method for out of distribution rejection of single instances \cite{geng2020recent}. Tracking the mean of recognition scores is a natural way to consider assessing if the input distribution has changed.  To design a reliability assessor, we use this, which is the simple estimate of a distribution, its mean. Using the observation that if the distribution changes to include more OOD samples, one would expect the mean SoftMax score to also be reduced. In this method, we collect the maximum value of SoftMax for each image in the batch and save it in a buffer, where buffer size is equal to batch size. Then, we compute the mean of the buffer. Finally, reliability can be assessed by comparing the mean with a calibrated mean value, e.g., a mean calibrated to optimize performance on a validation set with some fixed mixing ratio. Algorithm~\ref{a:meanSoftMax} in supplemental material summarizes this method.

\paragraph{Kullback-Leibler Divergence of Truncated Gaussian model of SoftMax Scores}
The mean of the maximum SoftMax only considers the first moment of the distribution. Therefore, if the distribution of a batch of images changes while the mean of distribution remains constant or changes a little, the first method fails. A more advanced method for change detection is using Kullback–Leibler divergence with a more advanced distributional model of the data. 

The Kullback-Leibler divergence is one of the most fundamental measures in information theory, which measures the relative entropy
\begin{equation}
\label{eq_KL}
\textup{KL} \;( P \lVert Q ) = \int_{-\infty}^{+\infty} p(x) \; \log(\frac{p(x)}{q(x)}) \; dx
\end{equation}
where $x$ is maximum SoftMax, $p(x)$ is probability density function of testing batch, and $q(x)$ is probability density function of training data set.

Making the classic Gaussian assumption for the distribution, i.e. letting $p(x) \sim \mathcal{N}(\mu,\,\sigma^2)$ and $q(x) \sim \mathcal{N}(m,\,s^2)$, we can derive the KL divergence measure as:
\begin{equation}
\label{eq_KL1} 
\textup{KL} \;( P \lVert Q ) = \log (\frac{s}{\sigma}) + \frac{ \sigma^2 + (\mu - m )^2}{2 \, s^2} - \frac{1}{2}
\end{equation}
Algorithm~\ref{a:KLSoftMax} in the supplemental material summarizes this method, which is
a direct application of KL divergence, as a measure between distribution of SoftMax values. Looking at the distribution in Fig.~\ref{fig_distribution}(a), one might note that the distibution is not at all symmetric and is bounded above by 1, which is the mode. 
Hence, it is  not well modeled by a Gaussian with simple mean and standard deviation of the raw data. The important insight here is to  consider this a truncated Guassian and use a moment-matching approach for approximation, such as \cite{kuss2005assessing}.  

\paragraph{KL divergence of Truncated Gaussian-model of EVM scores}
\label{subsec:KL_EVM}

It is, of course, natural to also consider combining the information theory-based KL divergence method with EVM-based probability estimates on each batch.
Algorithm~\ref{a:KLEVM} in the supplement provides the details.
Again, the distribution of scores shown in Fig.~\ref{fig_distribution}(b) is bounded, and this time with two disparate peaks, they seem even less Gaussian-like.
This is a novel combination of KL divergence and EVM scores and is included in the evaluation to help assess the information gain from EVM probabilities over SoftMax with exactly the same algorithm.

\paragraph{Fusion of KL divergence models}
\label{subsec:Bivariate_KL_fusion}
KL divergence of SoftMax and EVM are both viable methods with different error characteristics. Thus, it is also natural to ask if there is a way to fuse these algorithms. With KL divergence, an easy way to do this is to generalize to a bi-variate distributional model. We do this with a 2D  Gaussian model, using a full 2x2 co-variance model, and Algorithm~\ref{a:biKL} summarizes this method. Plots in the supplemental material show its performance, which is not significantly different. Thus, given the added cost we don't recommend it.

\paragraph{Open-world novelty detection with EVM}
\label{subsec:OND_EVM}

There are many ways to consider EVM probabilities, and given that the shape of data in Fig.~\ref{fig_distribution} is not apparently Gaussian, we sought to develop a method with a different distributional assumption. We develop a novel algorithm, we call open-world novelty detection (OND), which is designed to use only examples that have significant novelty, i.e., high probability of being OOD samples. Not only do we build from the extreme-value theory probabilities from EVM, but rather than looking at the mean or a Gaussian model of all data, we consider the mean of only those samples whose values probability of an input being from an unknown class exceed a threshold. We then combine that with a hypothesis test, which assumes a Bernoulli distribution of known and unknown inputs. This allows us to not just detect a distributional change but also if the change is improving or reducing reliability. 

From (\ref{eq_EVM_p}), we expect
\begin{equation}
\begin{cases}
1 \geq m(x_\textup{known}) \geq \tau \\ 
\tau > m(x_\textup{unknown}) \geq 0
\end{cases}
\end{equation}
By subtracting from 1,
\begin{equation}
\begin{cases}
1 - \tau \geq 1 - m(x_\textup{known}) \geq 0 \\ 
1 \geq 1 - m(x_\textup{unknown}) > 1 - \tau
\end{cases}
\end{equation}
Let's define
\begin{equation}
\begin{cases}
u(x) := 1 - m(x)\\ 
\delta := 1 - \tau
\end{cases}
\end{equation}
Then,
\begin{equation}
\begin{cases}
\delta \geq u(x_\textup{known}) \geq 0 \\ 
1 \geq u(x_\textup{unknown}) > \delta
\end{cases}
\end{equation}
By subtracting from $\Delta \in \mathbb{R}^+$
\begin{equation}
\begin{cases}
\delta - \Delta \geq u(x_\textup{known}) - \Delta \geq - \Delta \\ 
1 - \Delta \geq u(x_\textup{unknown}) - \Delta > \delta - \Delta 
\end{cases}
\end{equation}
If $1 \geq \Delta > \delta > 0$ 
\begin{equation}
\begin{cases}
\max \{ 0 , u(x_\textup{known}) - \Delta \} = 0 \\ 

1 - \Delta \geq \max \{ 0 , u(x_\textup{unknown}) - \Delta \} > 0
\end{cases}
\end{equation}
In batch mode with $N$ images
\begin{equation}
\label{eq_sum1}
\begin{cases}
 \sum \max \{ 0 , u(x_\textup{known}) - \Delta \} = 0 \\ 
N(1 - \Delta) \geq \sum \max \{ 0 , u(x_\textup{unknown}) - \Delta \} > 0
\end{cases}
\end{equation}

For a given domain, one might assume the mixture of known and OOD classes can be modeled by a Bernoulli distribution with probability of $\rho$. 
\begin{equation}
\label{eq_Bernoulli}
f(x) = 
\begin{cases}
\rho &, x \in \textup{OOD}\\ 
1 - \rho &, x \in \textup{Known classes}
\end{cases}
\end{equation}
By applying (\ref{eq_Bernoulli}) in (\ref{eq_sum1})
\begin{equation}
\rho N (1 - \Delta) \geq \sum \max \{ 0 , u(x) - \Delta \} > 0
\end{equation}
over time $\rho$ will change. If $\rho$ is constant or decreases, image classifier remains reliable. However, if it increases, the image classifier should be updated. 

We can find the error of this hypothesis as 
\begin{equation}
\varepsilon := \max \{0, ( \frac{1}{N} \sum \max \{ 0 , u(x) - \Delta \} ) - \rho (1 - \Delta) \}
\end{equation}
Finally, we propose the OND policy for automatic reliability assessment via thresholding on $\varepsilon$, see Algorithm~\ref
{a:ONDEVM} in the supplemental material. Note this algorithm has added parameters $\Delta and  \rho$, which need to be estimated on training/validation data. Although not evaluated herein, this algorithm has the operational advantage that it can actually be trivially tuned to target expected OOD ratios by changing $\rho$.

\section{Experimental Results}

For all experiments in this paper, we use an EfficientNet-B3 \cite{tan2019efficientnet} network that was trained to classify the ImageNet 2012 training data, with 79.3\% Top-1 accuracy on the 2012 validation set. Then, we extract features for images in the training set and train EVM with these extracted features.
In the evaluation, we consider the overall detection rate, on-time detection, and mean absolute error (see Eq~\ref{eq:mae}) as our primary metrics. We also show a metric called total accuracy. We start by considering the many batches in each test sequence and scoring each batch as $1.0$ if it is correctly labeled as Reliable before the change or correctly labeled Unreliable after the change, and scoring $0.0$ if incorrectly labeled. Total accuracy is the number of batches correctly labeled divided by the total number of batches. 

\begin{figure}[tb]
\centering
\subfloat[Total Detection Percentage]{\includegraphics[width=2.6in]{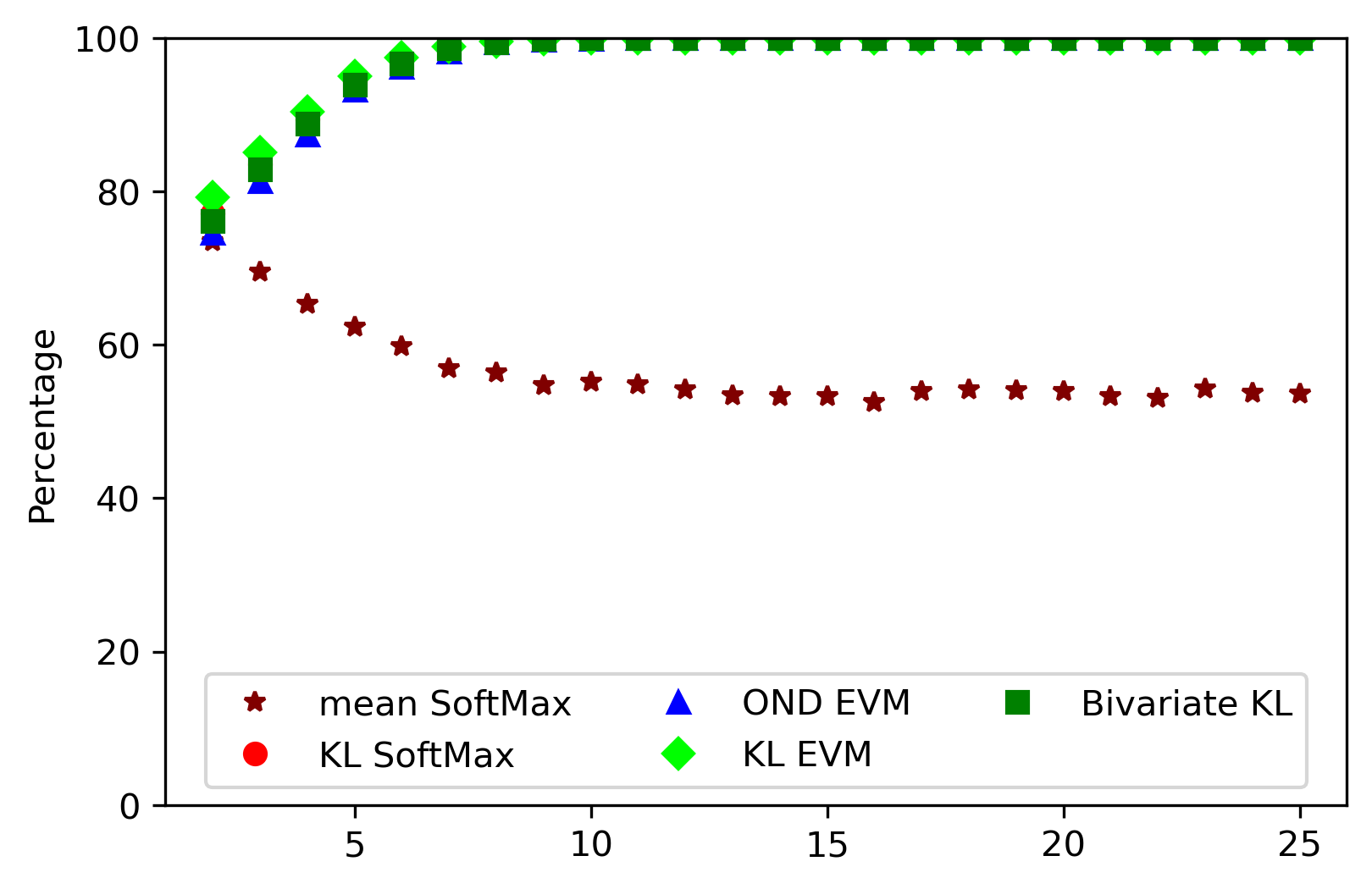}}\\
\subfloat[Percentage of on-time]{\includegraphics[width=2.6in]{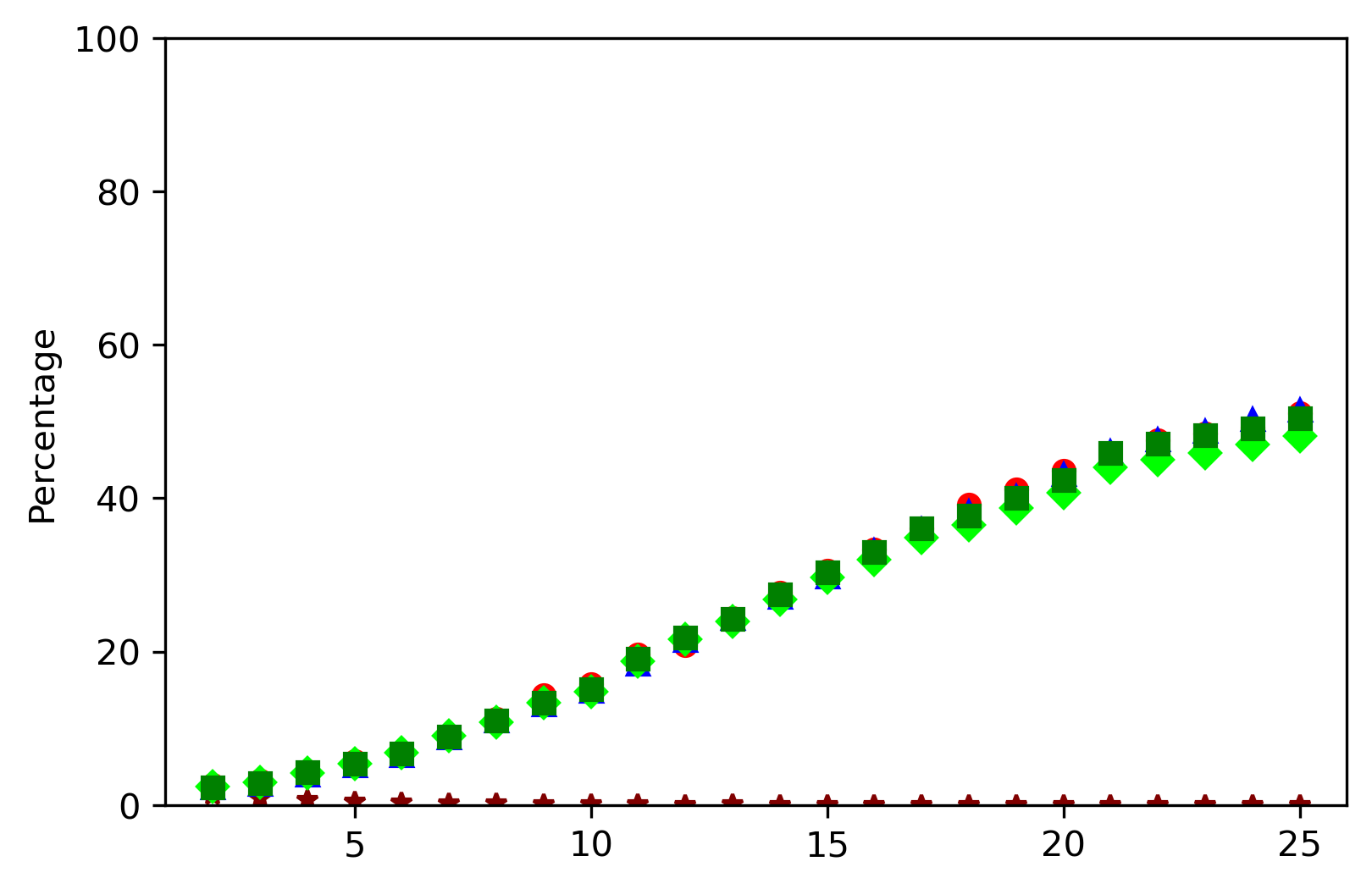}}\\
\subfloat[Mean Absolute Error]{\includegraphics[width=2.6in]{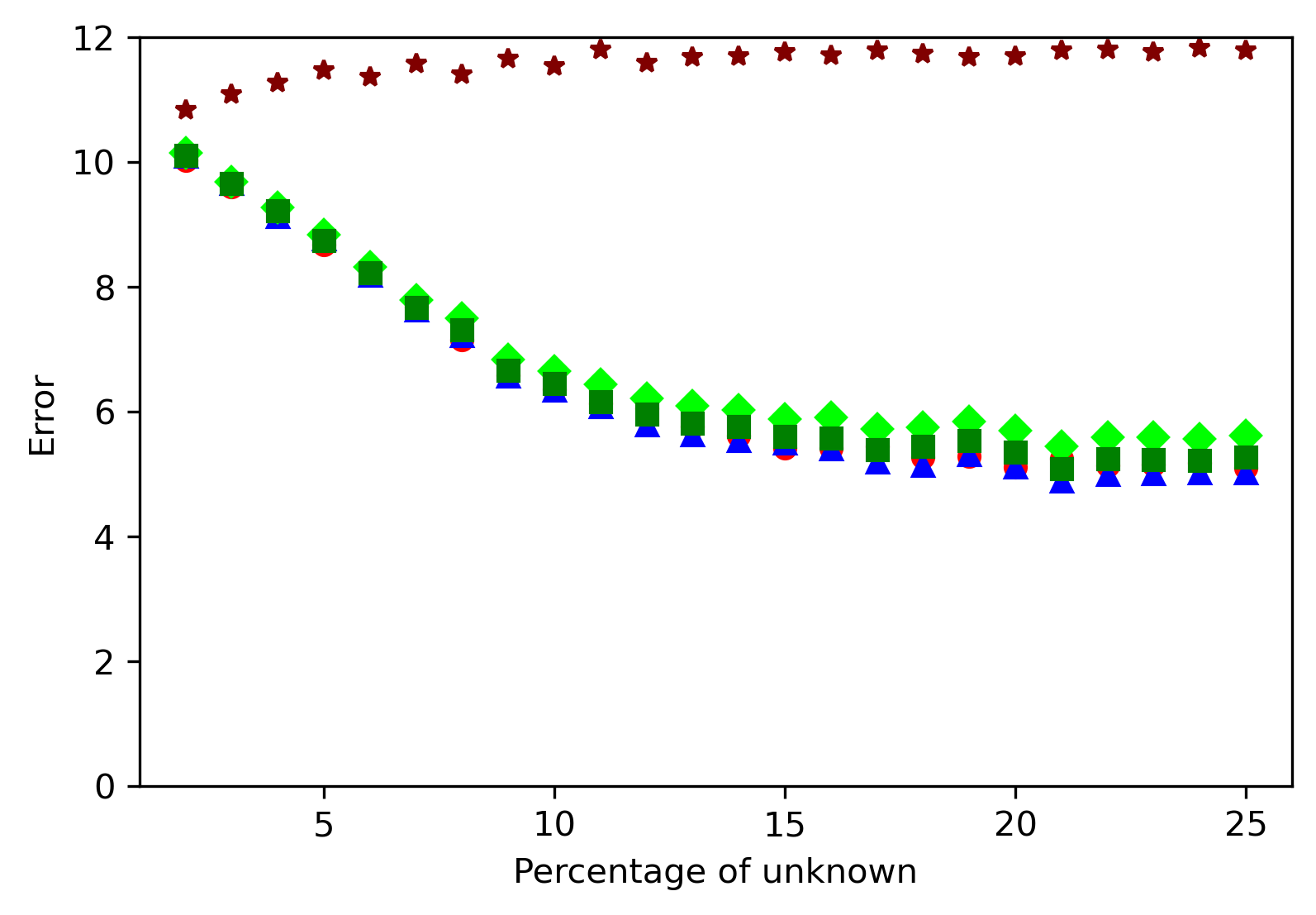}}
\caption{Performance of proposed policies when the threshold is selected to maximize the sum of on-time + late detection rates validation test with 2\% unknown.}
\label{fig_performance_over_unknown}
\end{figure}

\begin{figure}[tb]
\centering
\subfloat[True Detection, window size 20]{\includegraphics[width=2.6in]{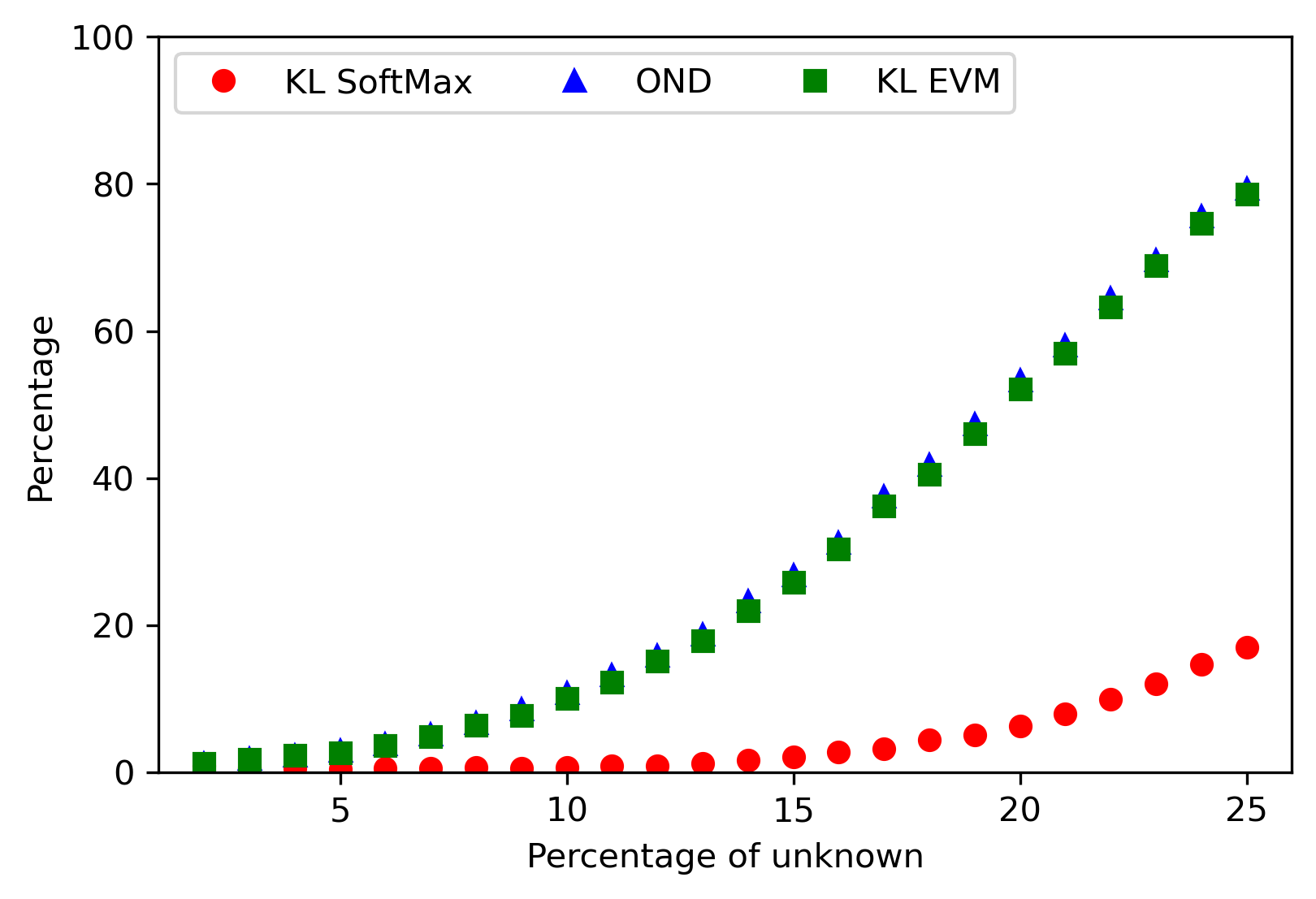}}\\
\subfloat[True Detection, window size  25]{\includegraphics[width=2.6in]{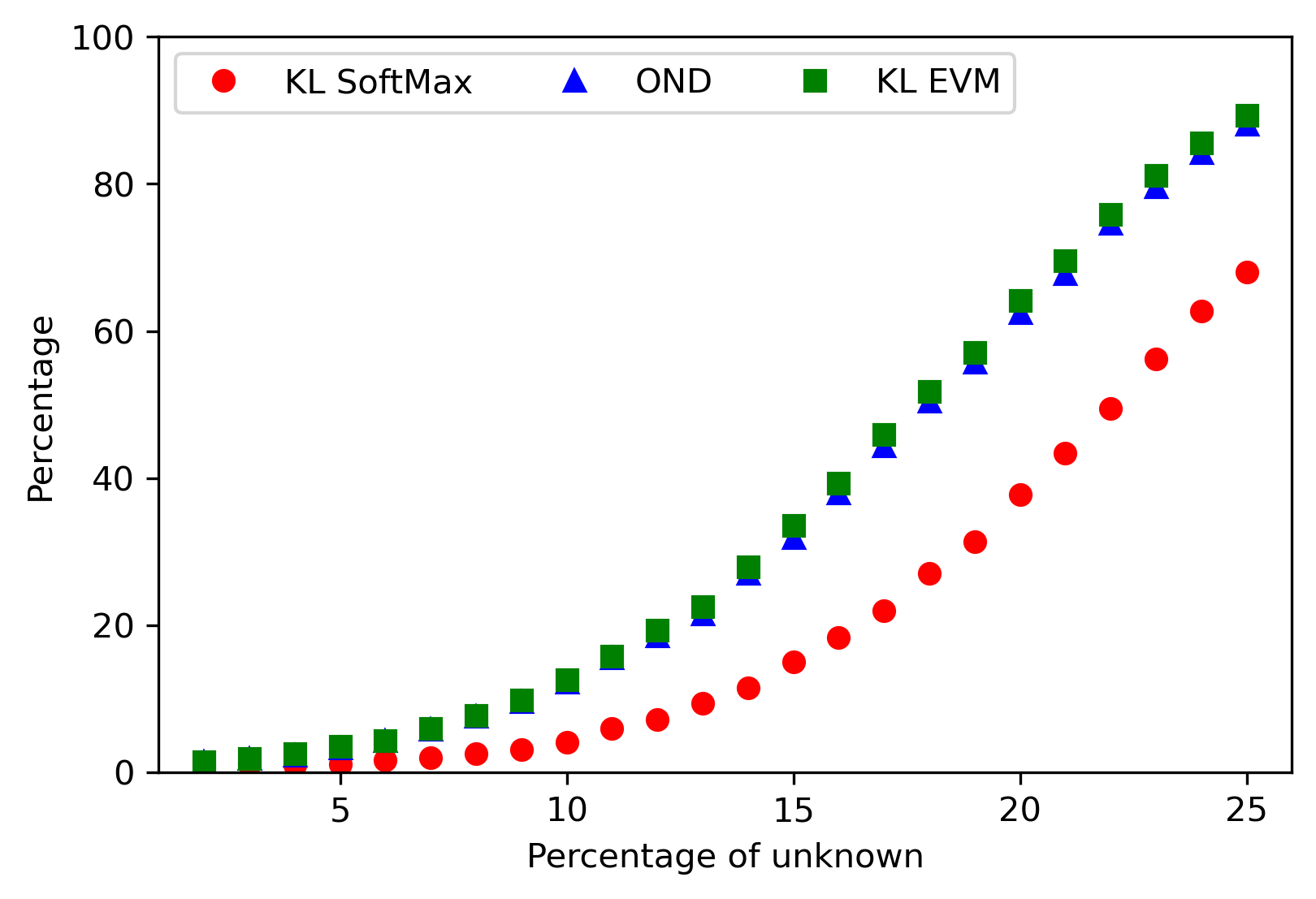}}\\
\caption{True detection performance with different sliding windows sizes for the proposed policies when the threshold is selected to limit false detection to at most 1\%.  We show true detection rate with is sum of on-time and late detection.}
\label{fig_performance_small_windows}
\end{figure}

In Fig.~\ref{fig_performance_over_unknown} we use window size of 100 and select each algorithms  threshold to maximize its true detection rate (sum of on-time and late detection) when the percentage of unknown (out-of-distribution) data is at 2\%. Operationally, users might determine the minimum variation they want to consider and the measure that is most important to them, then optimize that metric over the validation set to choose the parameter. For example, some users might prefer to maximize on-time detection or minimize false detection or mean absolute error.

Given these parameters, the performance of the various algorithms as the percentage of unknown data is increased, shown in Fig. \ref{fig_performance_over_unknown}.
For this setting, All of the distributional algorithms, the KL divergent of EVM or SoftMax (Sec~\ref{subsec:Bivariate_KL_fusion}) as well as the OND algoirthm~\ref{a:ONDEVM}, have similar good performance. The baseline mean of SoftMax-based  Algorithm~\ref{a:meanSoftMax} has by far the lowest detection rate, total accuracy, and on-time.

\begin{figure*}[t]
\centering
\subfloat[False Detection Percentage]{\includegraphics[width=.33\linewidth]{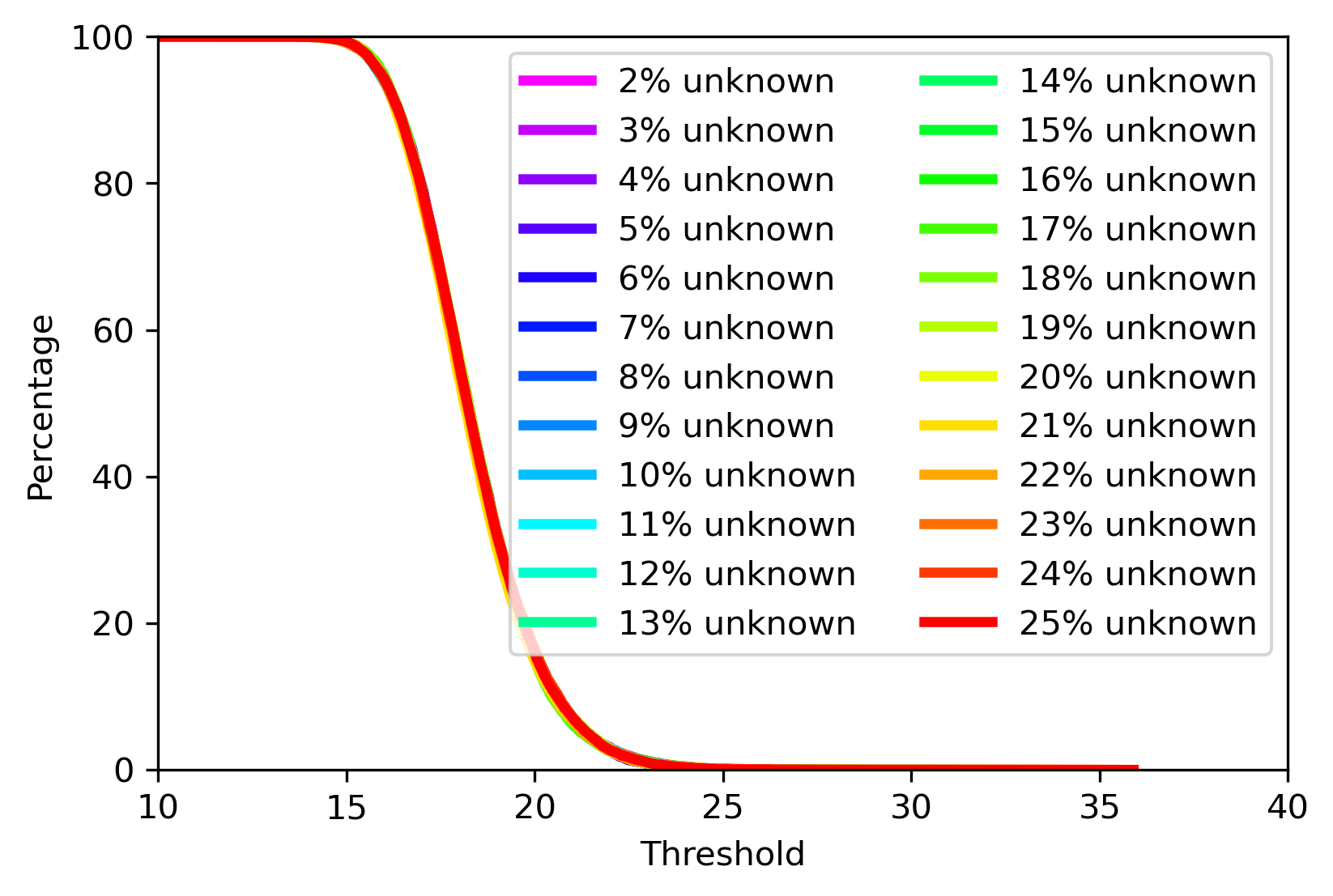}}
\subfloat[Total Accuracy]{\includegraphics[width=.33\linewidth]{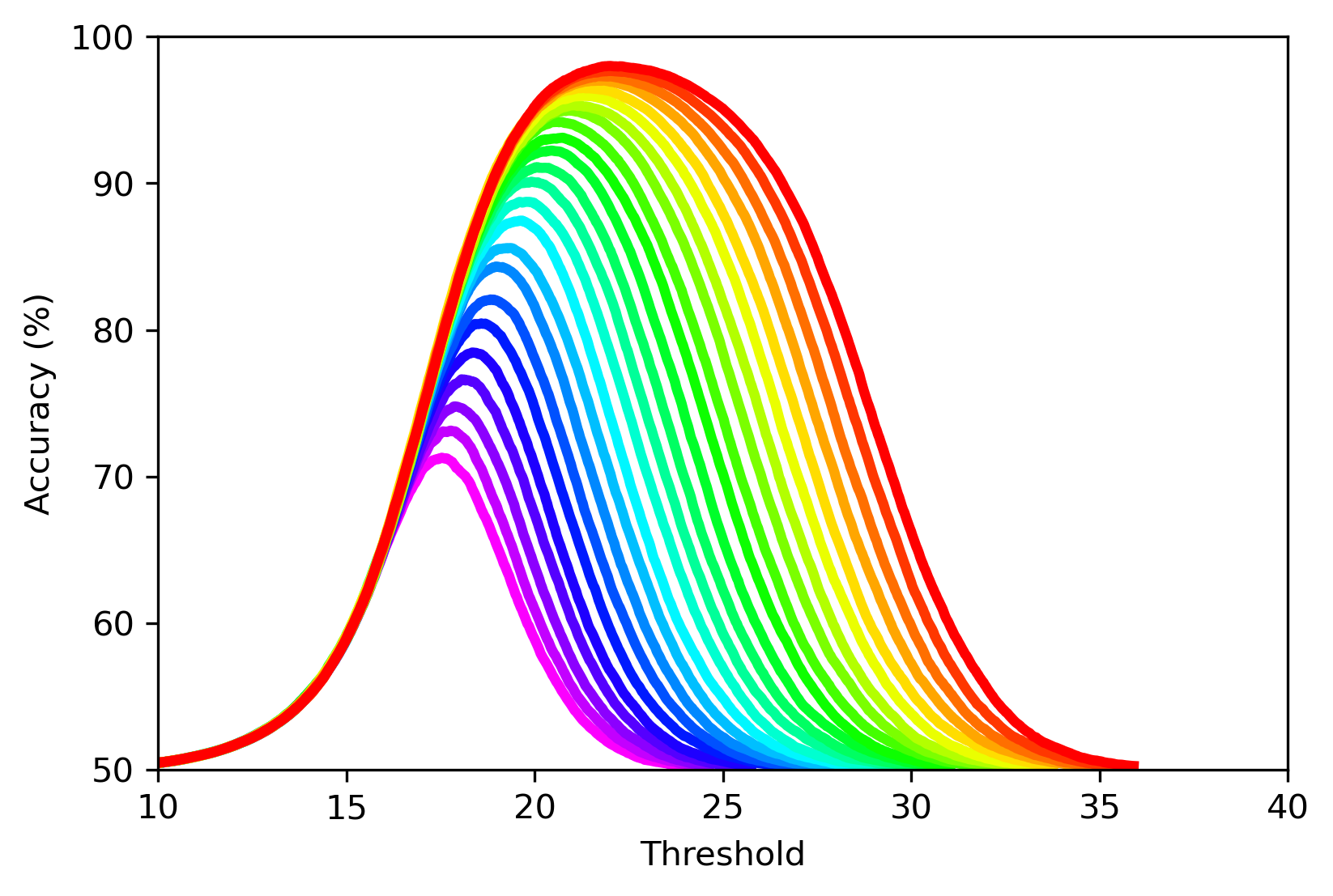}}
\subfloat[Total Detection Percentage]{\includegraphics[width=.33\linewidth]{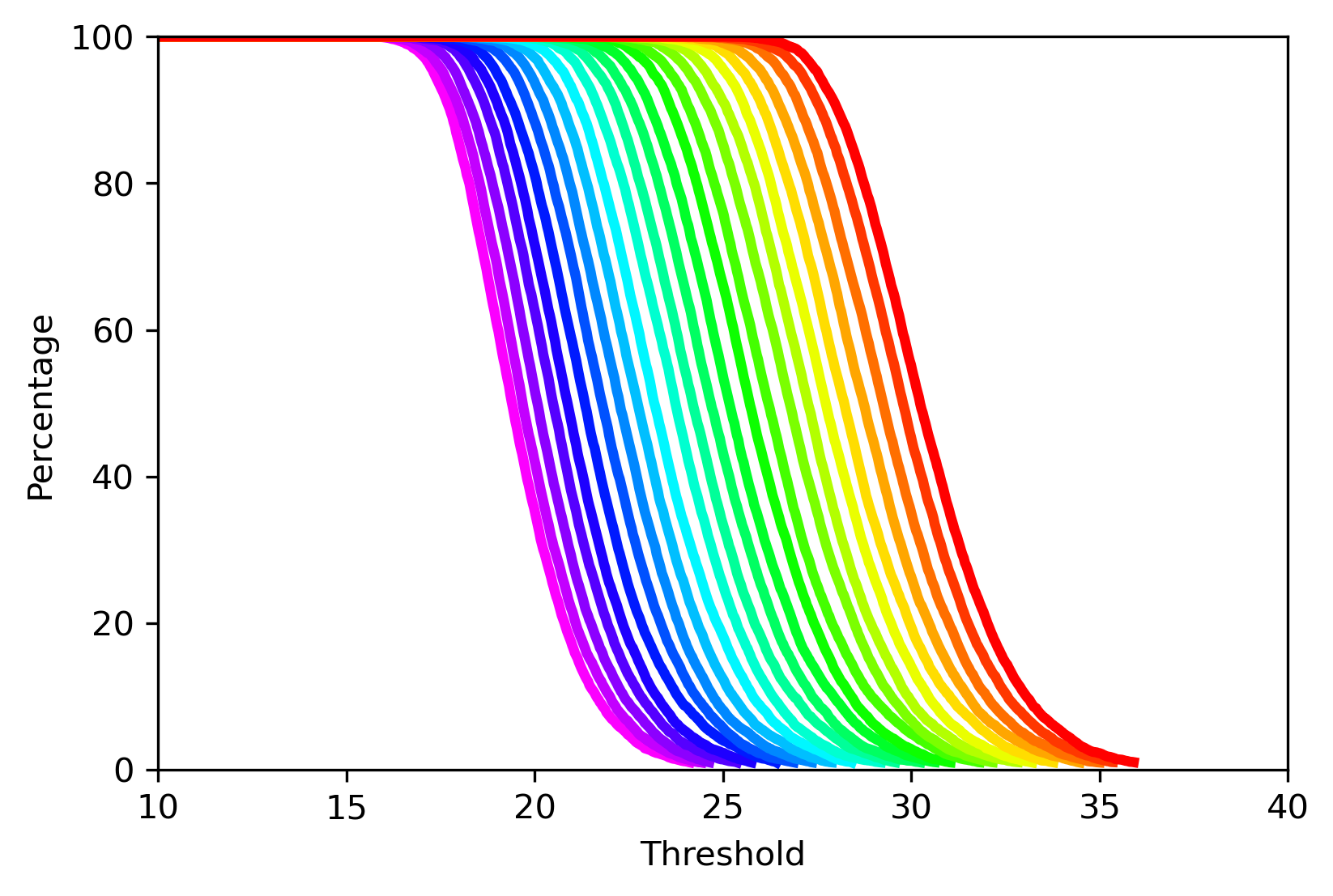}}\\
\subfloat[On-time Percentage]{\includegraphics[width=.33\linewidth]{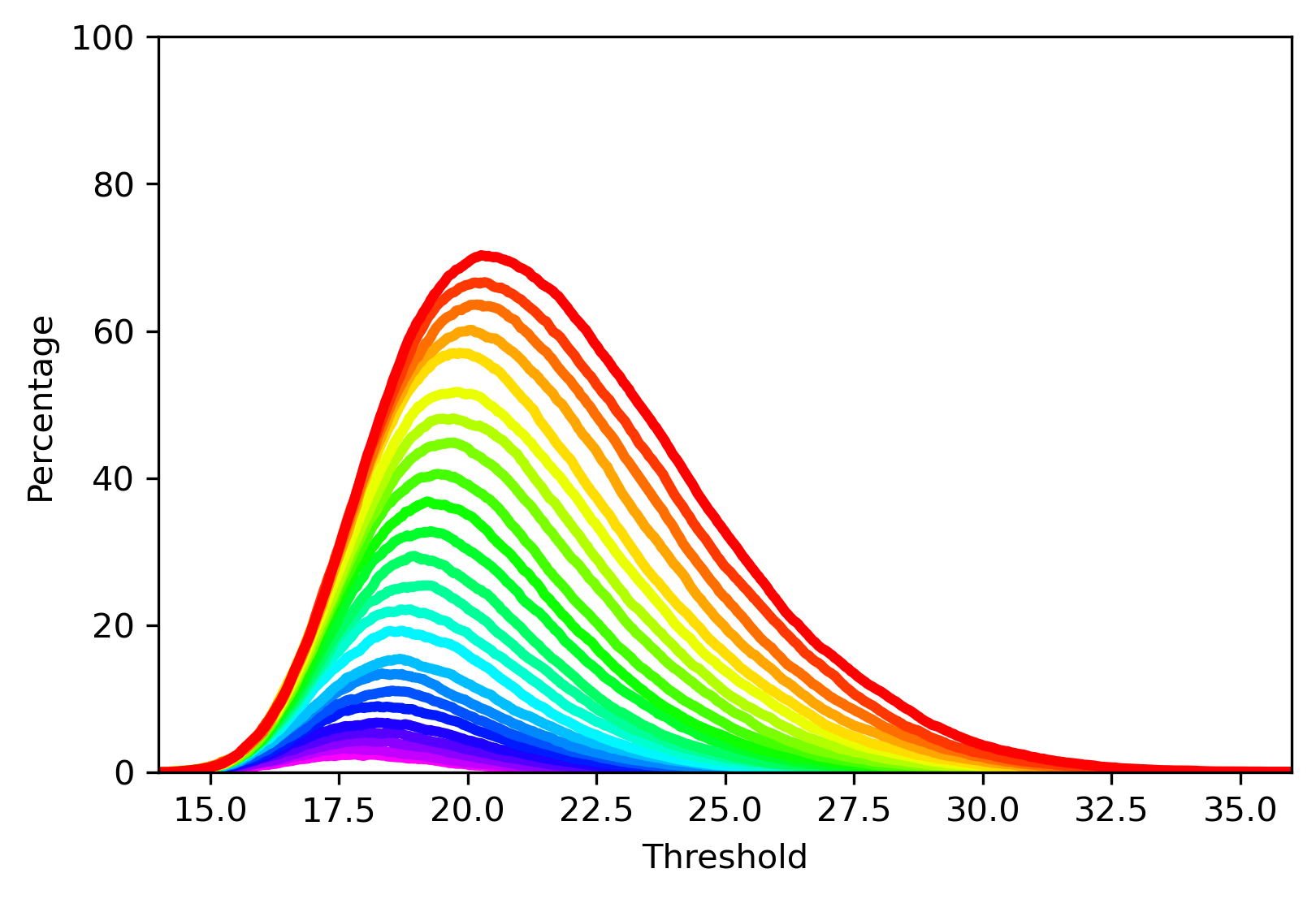}}
\subfloat[Late Percentage]{\includegraphics[width=.33\linewidth]{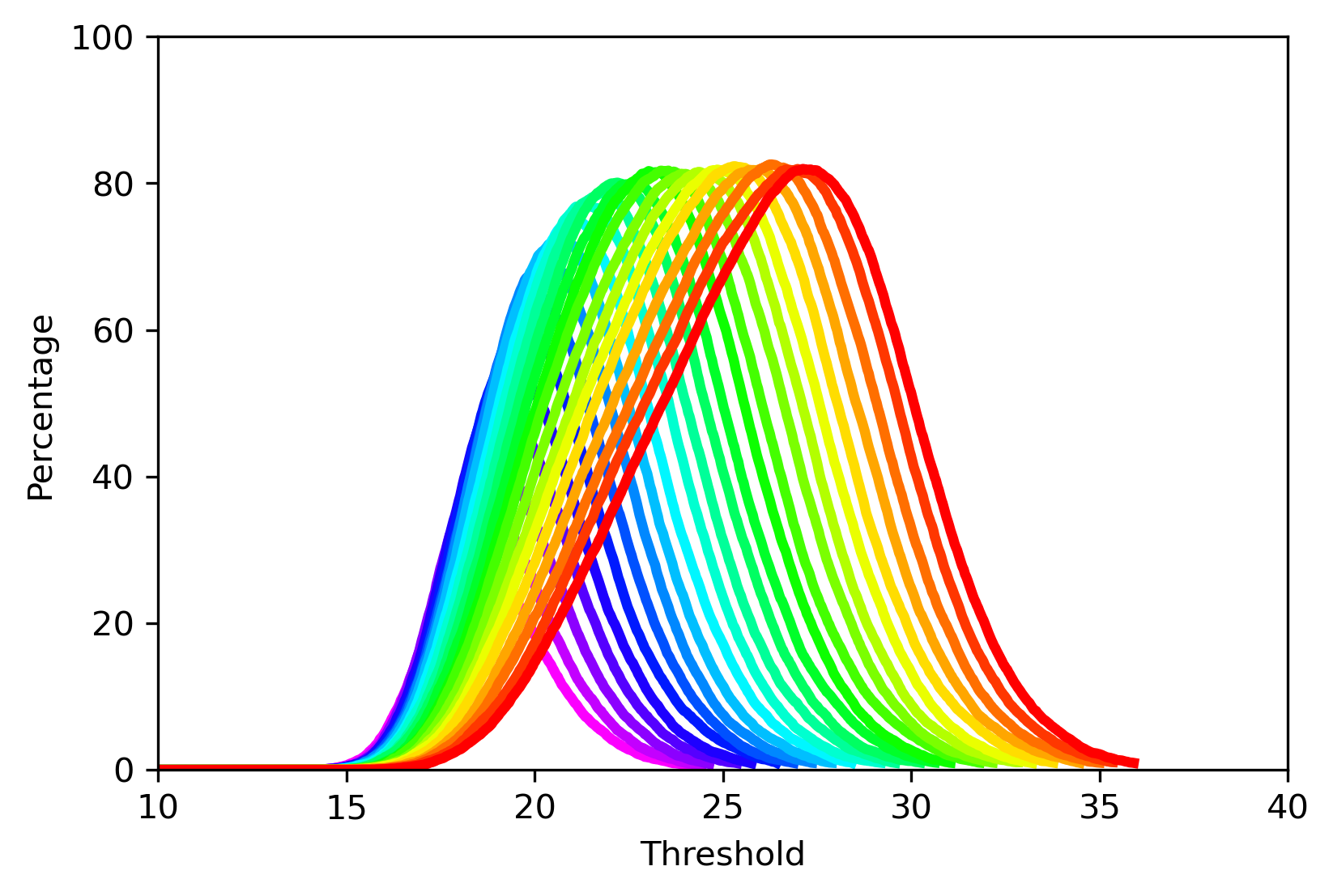}}
\subfloat[Mean Absolute Error]{\includegraphics[width=.33\linewidth]{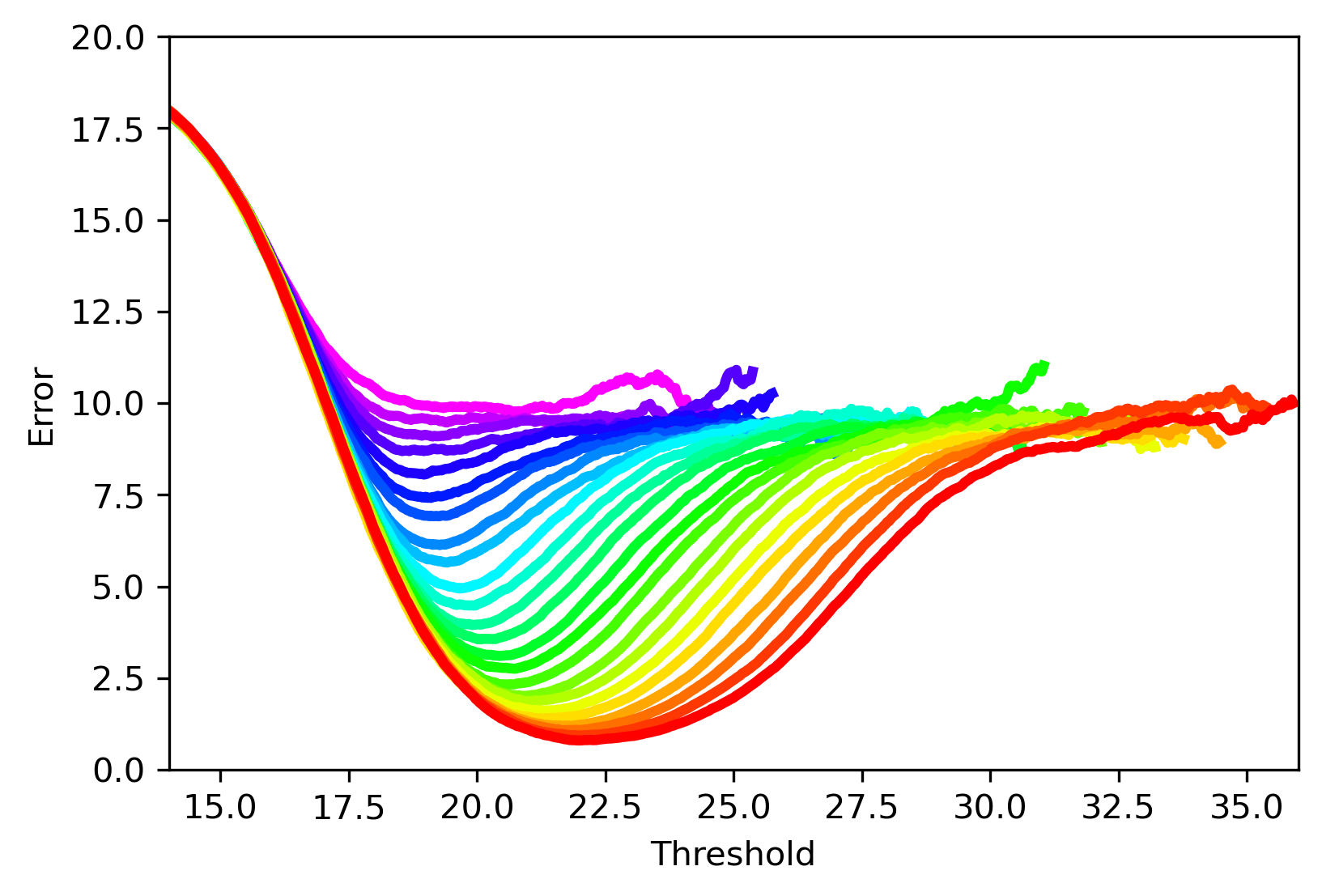}}
\caption{Ablation study on performance of Bivariate KL Fusion Algorithm Section~\ref{subsec:Bivariate_KL_fusion} when the threshold changes. 
As the percentage of unknown increases, the breadth of good algorithm performance
increases, and the peak performance of moves slightly to the
right (to higher thresholds).}
\label{fig_performance_over_threshold}
\end{figure*}

In the remaining plots in the paper, we are doing ablation studies.  Such ablation studies allow us to understand the deeper aspects of performance.
In Fig.~\ref{fig_performance_small_windows}, we  use batch/window size of 20 and 25 and select each algorithms  threshold to limit false detctions to at most 1\% when the percentage of unknown (out-of-distribution) data is at 2\%.  With smaller windows the difference between KL EVM and KL SoftMax become signficant, while KL EVM and OND are similar showing that the difference is the superior EVM features for detection of OOD data.

Next let us  consider peak performance, which we shall see as a function of the percentage of OOD/Unknown data, which is unknowable.
This ablation study is to determine the impact and sensitivity of thresholds, Fig.
\ref{fig_performance_over_threshold} demonstrates the effect of threshold on performance Algorithm~\ref{a:biKL} in
different percentages of unknown.

\section{Discussion}

Fig. \ref{fig_distribution} (on the first page) shows the distribution of the magnitude of the maximum probability of SofMax and EVM on known validation and OOD validation sets. 
There are few known images with low predicted probability, and for SoftMax, many unknowns with high probability. Therefore, for any given threshold, a significant portion of the unknown images will be classified as known. 
Thus, with neither of these base models, is it possible to design a reliability assessor without failures. 
This also makes it clear why very simple naive algorithms, such as thresholding on SoftMax and retraining when images with low scores are detected, are useless -- because the network is not perfect, and even training images will fail such a simple test, and the system will always be training. This is part of what makes this problem challenging and interesting -- the algorithms must account for the imperfect nature of the core classification algorithm as well as handling the open-set nature of the problem while detecting the change in OOD. 

The current EVM has three important parameters: cover threshold, tail size, and distance multiplier. We use 0.7 and 33998 for cover threshold and tail size, the same as the first EVM paper \cite{rudd2017extreme}. The EVM paper did not discuss a distance multiplier, seeming to use an implicit 0.5. We try 0.40, 0.45, 0.5, 0.75, and 1.0 for the distance multiplier. Among them, 0.45 demonstrates the best separation between known validation and unknown validation sets.

After training EVM, the threshold $\Delta$ must be selected for KL EVM Algorithm~\ref{a:KLEVM}. Because the cover threshold is 0.7, the $\Delta$ must be greater than 0.3 and less than 0.7. This data can be optimized based on the validation data set. However, with such a small region of allowed values, we decide to choose the middle, i.e., 0.5, to avoid bias. There is not any consensus or common sense to determine thresholds exactly, especially given the inherent trade-off between false detection and failure. As the percentage of failure decreases, the false detection increases and vice versa. Studying all possible methods to tune the thresholds is beyond the scope of this paper. One approach is to select a threshold slightly above the signal to avoid early detection. This approach in practice can be found by creating a validation set and updating it in a timely manner according to the need and scene of the images in a real test.

We note that our KL divergence models are assuming a truncated Gaussian distribution, which is only an approximation to the shapes of the distributions shown in Fig.~\ref{fig_distribution}, and the quality of the approximation will be impacted by mixing and by window sizes. While only an approximation, the KL divergence does much better than the simple detection of a shift in the mean. The OND algorithm makes very different assumptions using a Bernoulli model. The nearly equal performance of KL divergence assuming truncated Gaussian and OND assuming Bernoulli suggest the overall detection is relatively stable and not strongly dependent on either set of assumptions. 

A good algorithm must detect the change to unreliability even with a small input distribution shift of only a few percent and it does. However, our experiments show that even with a large distribution shift, i.e., 25\% OOD data, algorithms fail to always detect the shift on-time.  The main reason is that features of unknown images can overlap with known images, and, therefore algorithms cannot sense the shift. In a few splits, Kullback–Leibler divergence of SoftMax (baseline) works somewhat better than algorithms that use EVM information. However, when the shift is larger, the algorithms that use EVM information outperform the algorithms that use only SoftMax.  For lower percentages of unknown KL EVM does slightly better in total detection, and for larger percentages OND has slightly lower mean absolute error of detection, but neither are very significant differences.  

The sensitivity and accuracy of all presented methods increases as batch size increases.  However, the batch size is proportional to the inverse of speed. If the batch size goes to infinity, the speed of detection goes to zero, which is not practical. Thus, in the trade-off between accuracy and speed, we should find a conformation to reduce the batch size while the distribution approximately is a Gaussian. For the ImageNet data, we found that a batch size of 100 is a good balance for both and EVM is reasonably stable down to windows of 20, but KL on SoftMax requires larger batches.

We note that all experiments herein were with ImageNet classifiers which have 1000 classes. For smaller problems, there may be a greater   issue of OOD classes folding on-top of known classes(see \cite{dhamija2018reducing}), and hence the difference between SoftMax and EVM based detection may be more apparent for problems with smaller number of classes.

\section{Conclusions}

Both standard closed-set and more recent open-set classifiers face an open-world, and errors can increase when there is a change in the distribution of the frequency of unknown inputs. While open-set image classifiers can map images to a set of predefined classes and simultaneously reject images that do not belong to the known classes, they cannot predict their own overall reliability when there is a shift in the OOD distribution of the world. Currently, one must allocate human resources to monitor the quality of closed or open-set image classifiers. A proper automatic reliability assessment policy can reduce the cost by decreasing human operators and simultaneously increasing user satisfaction. 

In this paper, we formalized the open-world reliability assessment problem and proposed multiple automatic reliability assessment policies (Algorithms~\ref{a:KLSoftMax}, \ref{a:ONDEVM},  \ref{a:KLEVM}, and \ref{a:biKL}) that use only score distributional data. Algorithm~\ref{a:ONDEVM} uses the maximum of the mean of thresholded maximum EVM class probability to determine reliability. This algorithm ignores the higher order moments of distribution for simplicity. Algorithm \ref{a:KLEVM} uses Gaussian Model of score distributions and Kullback–Leibler divergence of maximum EVM class probabilities. We also evaluate, Algorithm~\ref{a:KLSoftMax}, that uses the Kullback–Leibler divergence on a Gaussian Model of SoftMax scores, and show it also provides an effective open-world reliability assessment.

We used the mean of SoftMax as the baseline, and the new algorithms are significantly better. We used ImageNet 2012 for known images and non-overlapping ImageNet 2010 for unknown images. As in any detection problem, there is an inherent trade-off between false or early-detection and true detection rates.  If we tune the threshold, so there are zero-false detection, then even the best algorithm fails to detect in 5.68\% of tests. This is because features of unknown images can overlap on known images, and therefore, algorithms cannot sense the shift. In future works, we will investigate extreme value theory rather than Gaussian assumptions to design a better automatic reliability assessment policy.

{\small
\bibliographystyle{ieee_fullname}
\bibliography{ARSbib}
}

\clearpage

 \section*{Supplemental Material}

In the supplemental material
we also present a plot that combines on-time and total-detection rates. We also present plots obtained when we optimize to total-accuracy rather than on-time performance. Finally, we present pseudo-code for all algorithms considered in the paper.

\subsection*{Bivariate KL fusion}
If 
\begin{equation}
p(x) \sim \mathcal{N}(
\begin{bmatrix}
\mu_1\\
\mu_2
\end{bmatrix}
,\,\begin{bmatrix}
\sigma_1^2 & \rho \sigma_1 \sigma_2 \\
\rho \sigma_1 \sigma_2 & \sigma_2^2
\end{bmatrix}),
\end{equation}
\begin{equation}
q(x) \sim \mathcal{N}(\begin
{bmatrix}
m_1\\
m_2
\end{bmatrix}
,\,\begin{bmatrix}
s_1^2 & r s_1 s_2 \\
r s_1 s_2 & s_2^2
\end{bmatrix})
\end{equation}
Then, (\ref{eq_KL}) can be written as
\begin{multline}
\label{eq_KL2_full}
\textup{KL} \;( P \lVert Q ) = \log (\frac{s_1 s_2 \sqrt{1-r^2}}{\sigma_1 \sigma_2 \sqrt{1-\rho^2}}) + \frac{1}{2(1-r^2)} ( \\
\frac{(\mu_1 - m_1)^2 + (\sigma_1 - s_1)^2}{s_1^2} + \frac{(\mu_2 - m_2)^2 + (\sigma_2 - s_2)^2}{s_2^2} \\
- 2r \frac{(\mu_1 - m_1)(\mu_2 - m_2) + \rho \sigma_1 \sigma_2 - r s_1 s_2}{s_1 s_2} )
\end{multline}

\subsection*{Plots}

\begin{figure}[h]
\centering
\includegraphics[width=2.8in]{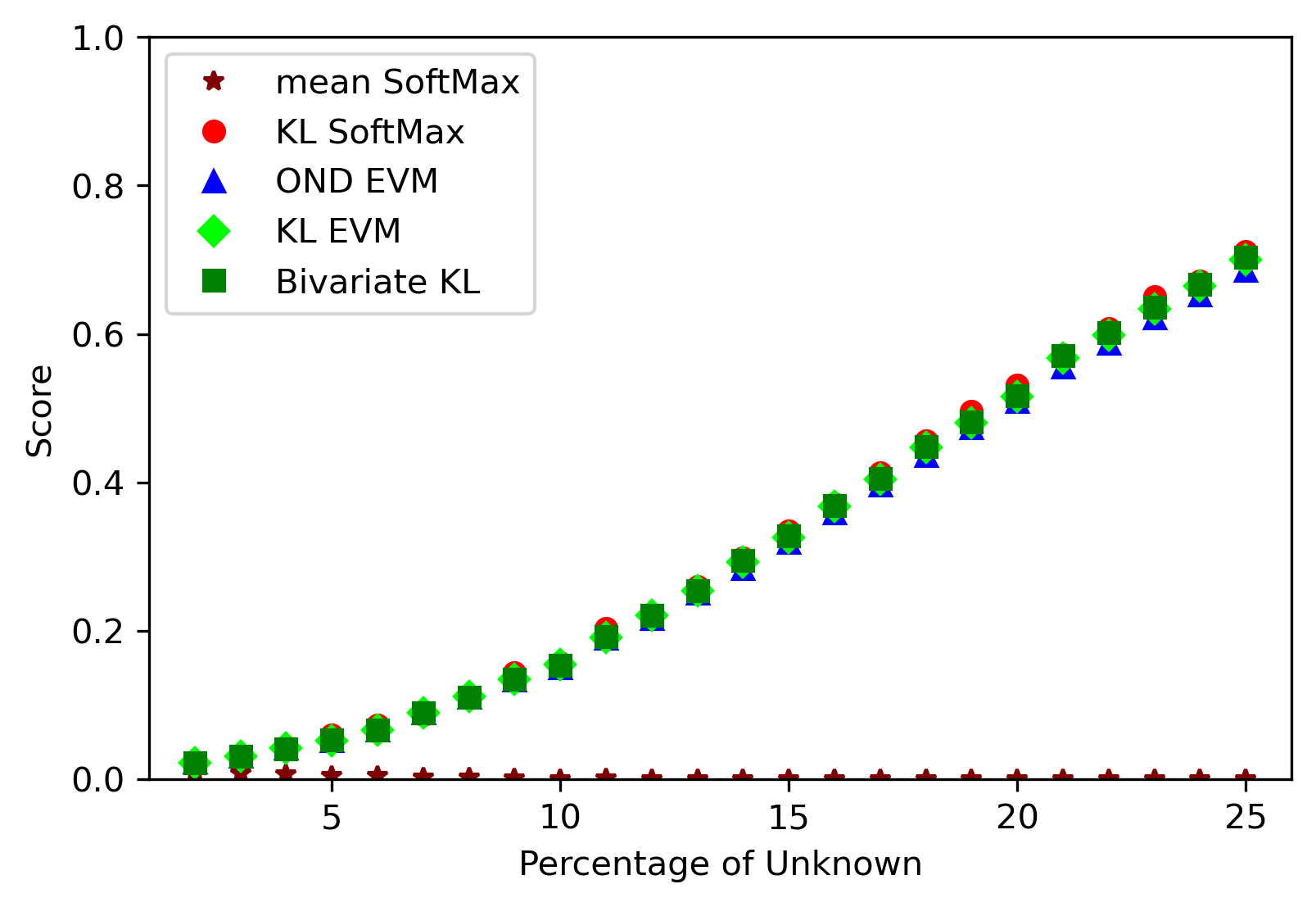}
\caption{Reliability Score of proposed algorithms when the best threshold for each point is selected.}
\label{fig_score}
\end{figure}

To facilitate overall comparison We can consider a combined reliability score defined as $on\_time\_rate * total\_detection\_rate$. Fig. \ref{fig_score} demonstrate the score of proposed algorithms when each point has a different threshold to compare the best performance of each algorithm in different percentage of unknown. The score is computed as the maximum multiplication of on-times and total detected ratio over all possible thresholds. We can see that the algorithm performs well for  distributions of either EVM data or SoftMax value for decisions.

\begin{figure}[htb!]
\centering
\subfloat[Total Accuracy]{\includegraphics[width=2.5in]{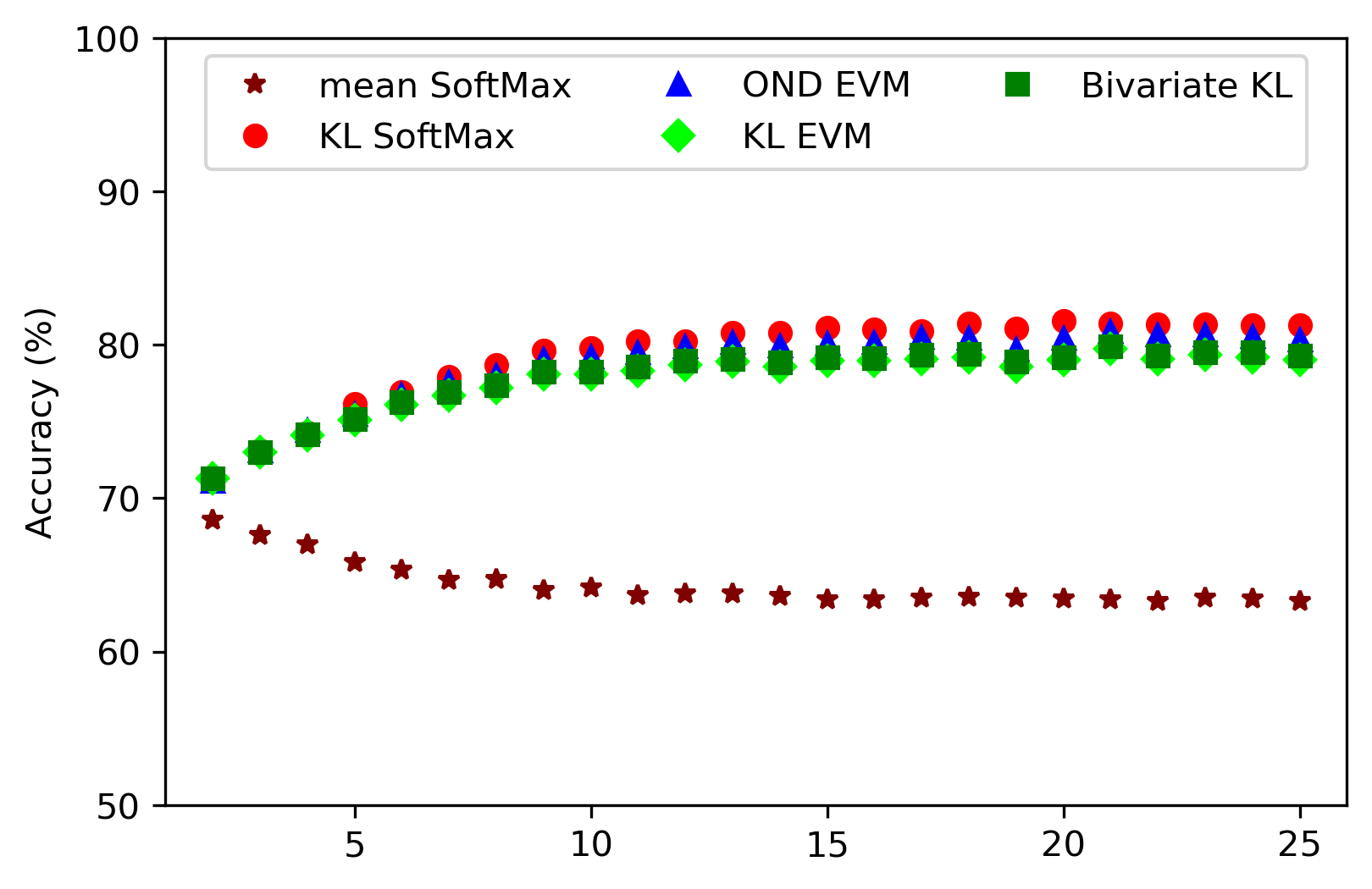}}\\
\subfloat[Total Detection Percentage]{\includegraphics[width=2.5in]{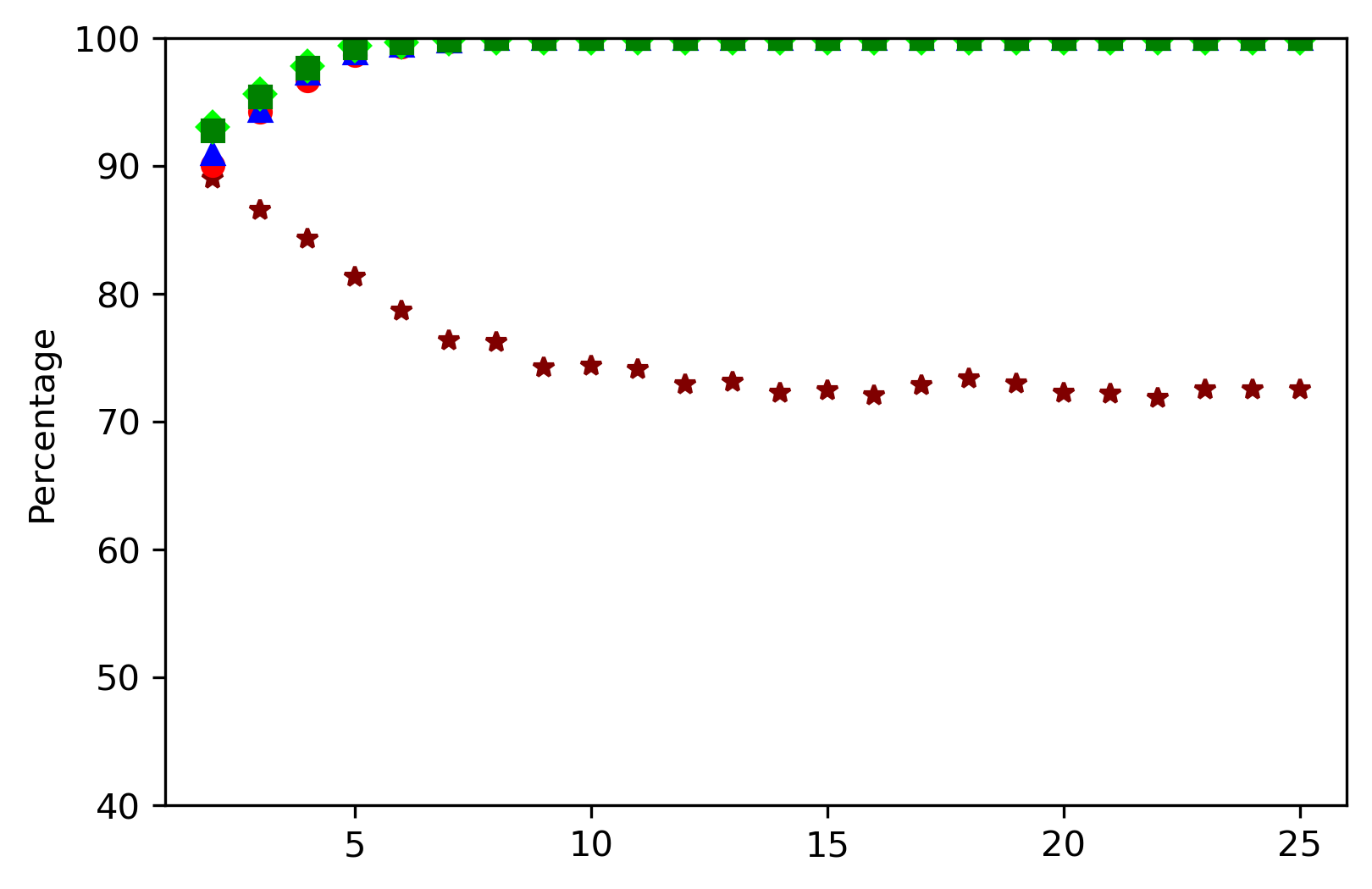}}\\
\subfloat[Percentage of on-time]{\includegraphics[width=2.5in]{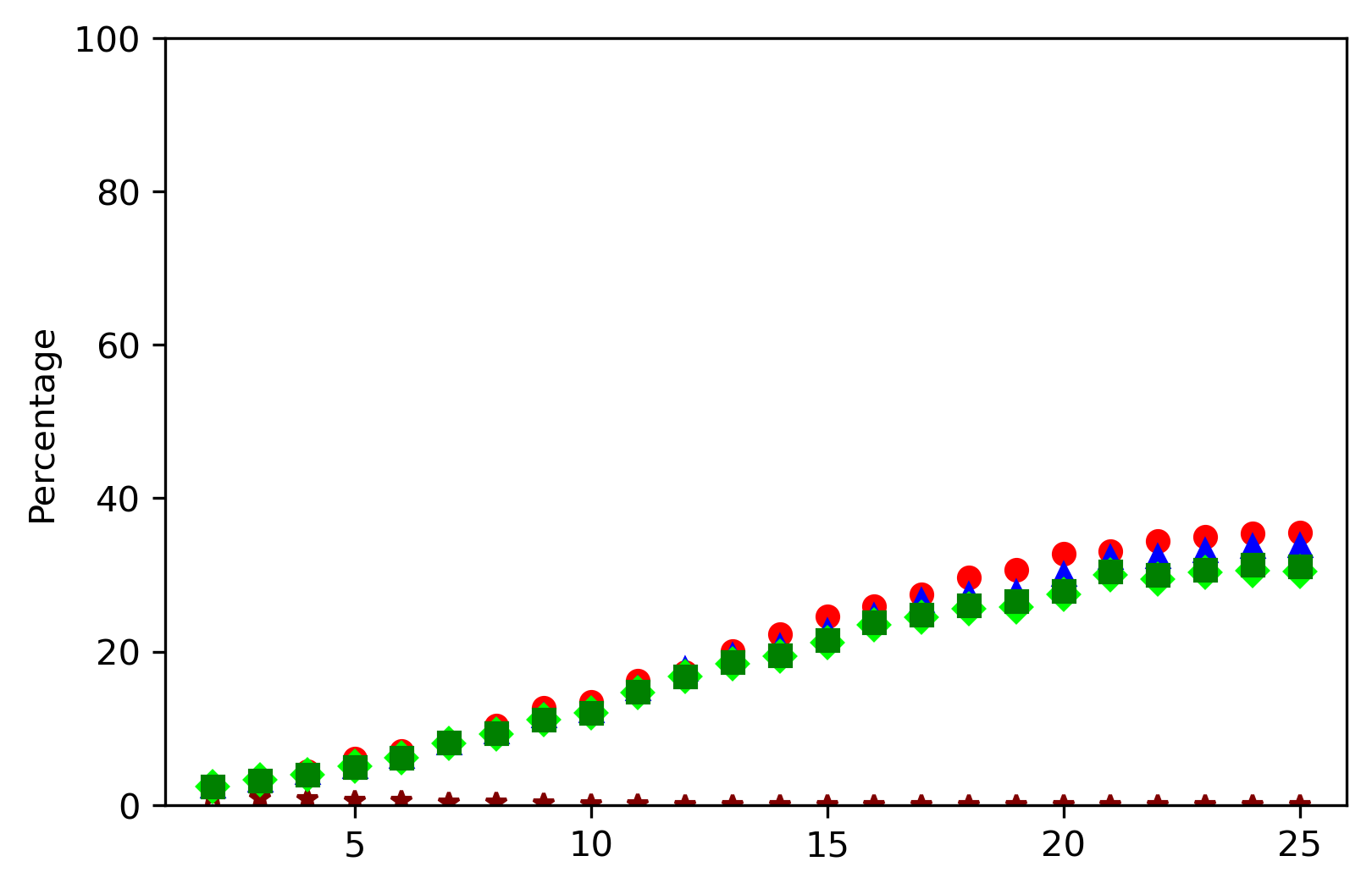}}\\
\subfloat[Mean Absolute Error]{\includegraphics[width=2.5in]{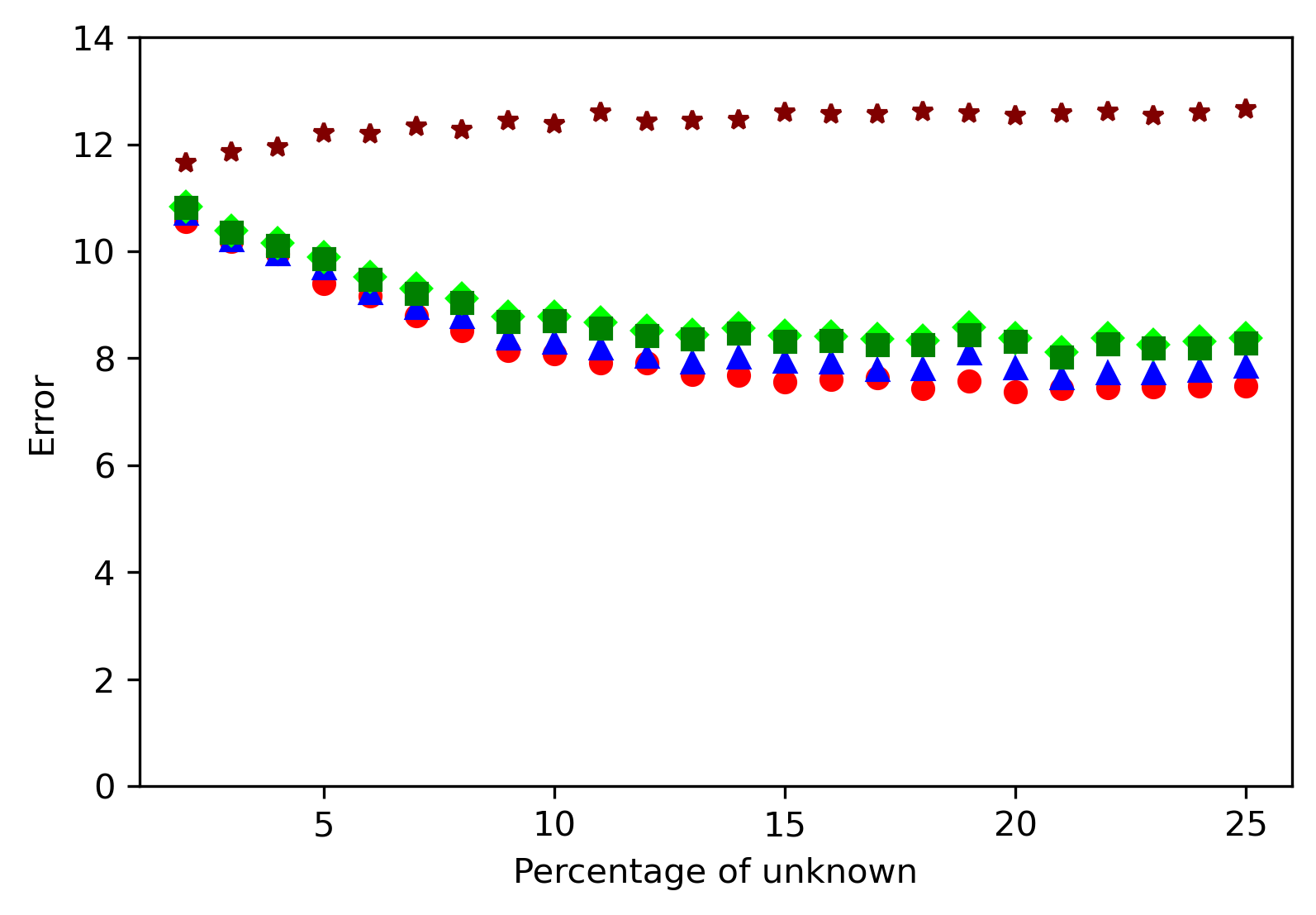}}
\caption{Performance of proposed policies when the threshold is selected to maximize the total accuracy validation test with 2\% unknown. Compare with Fig~\protect\ref{fig_performance_over_unknown} in main paper.}
\label{fig_performance_over_unknown_total_accuracy}
\end{figure}

\begin{figure}[htb!]
\centering
\subfloat[Total Accuracy]{\includegraphics[width=2.5in]{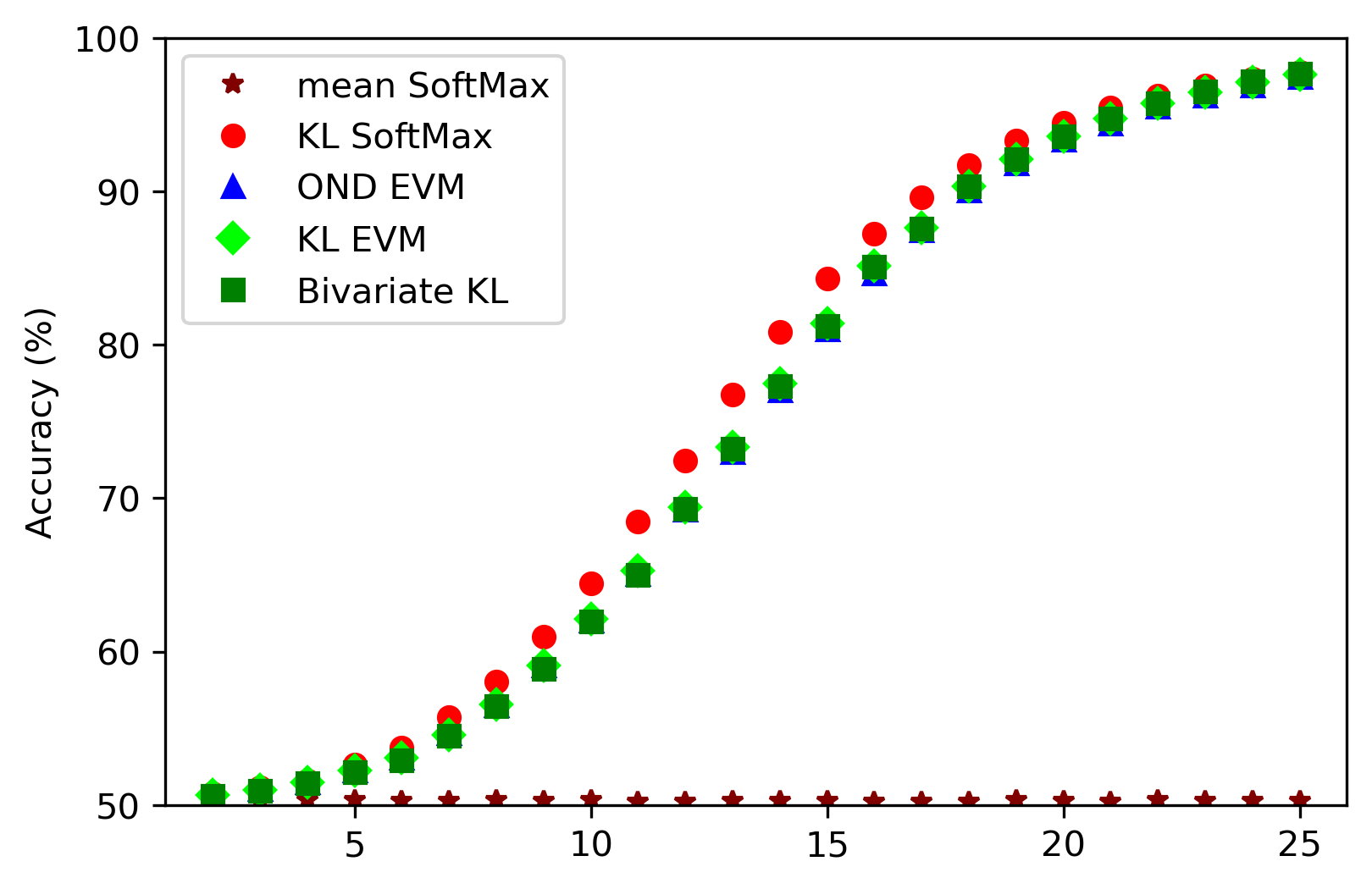}}\\
\subfloat[Total Detection Percentage]{\includegraphics[width=2.5in]{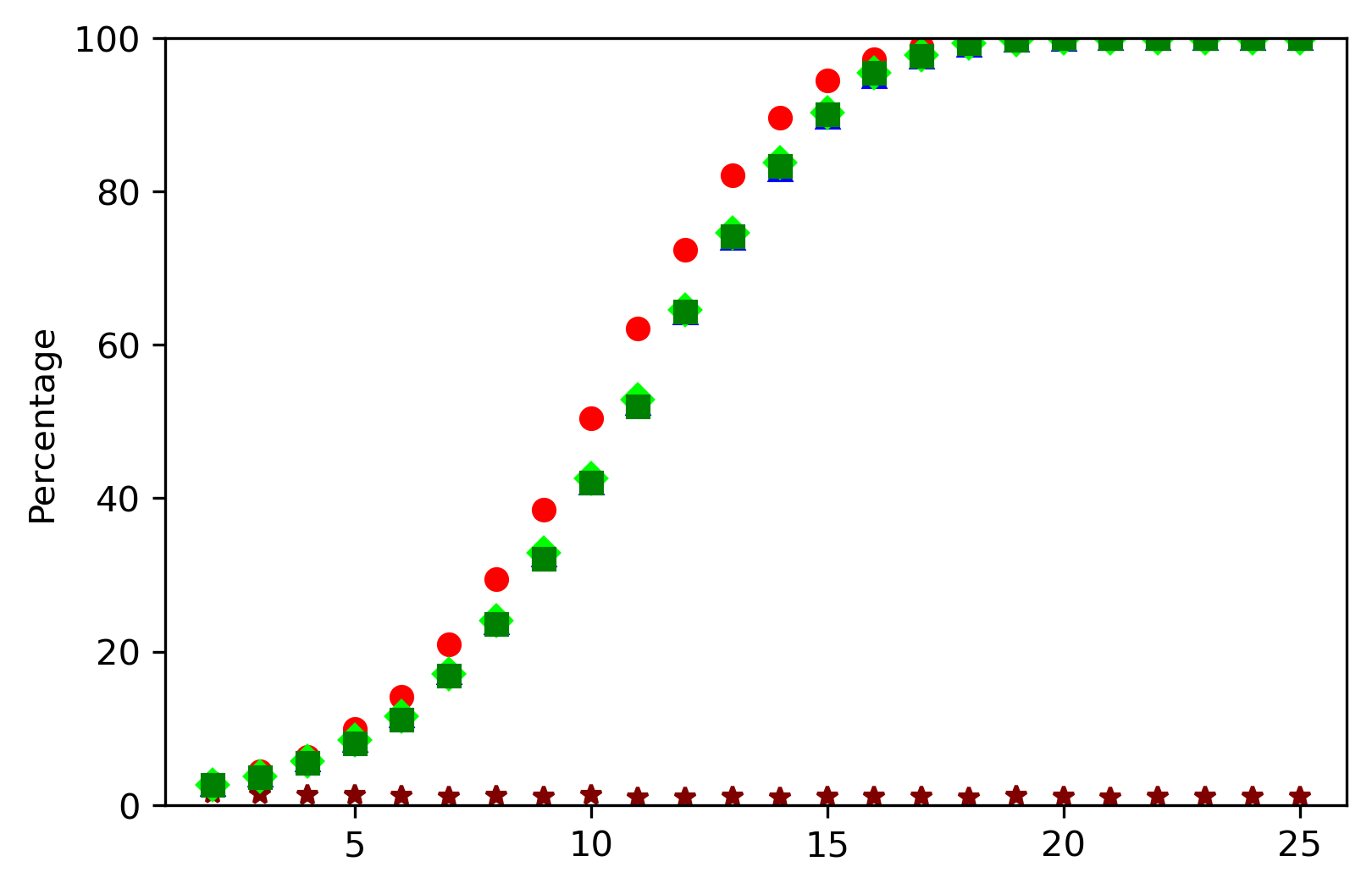}}\\
\subfloat[Percentage of on-time]{\includegraphics[width=2.5in]{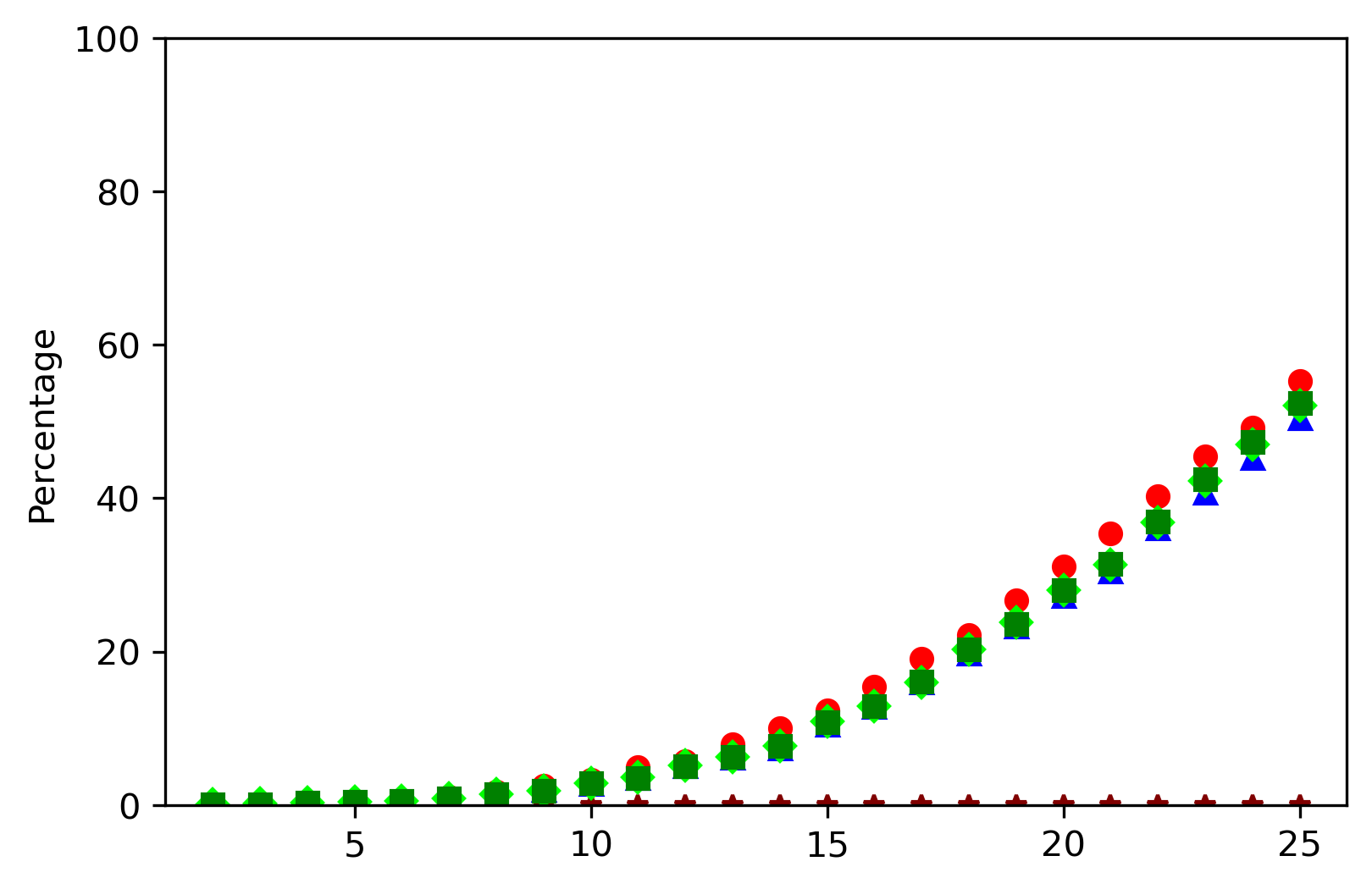}}\\
\subfloat[Mean Absolute Error]{\includegraphics[width=2.5in]{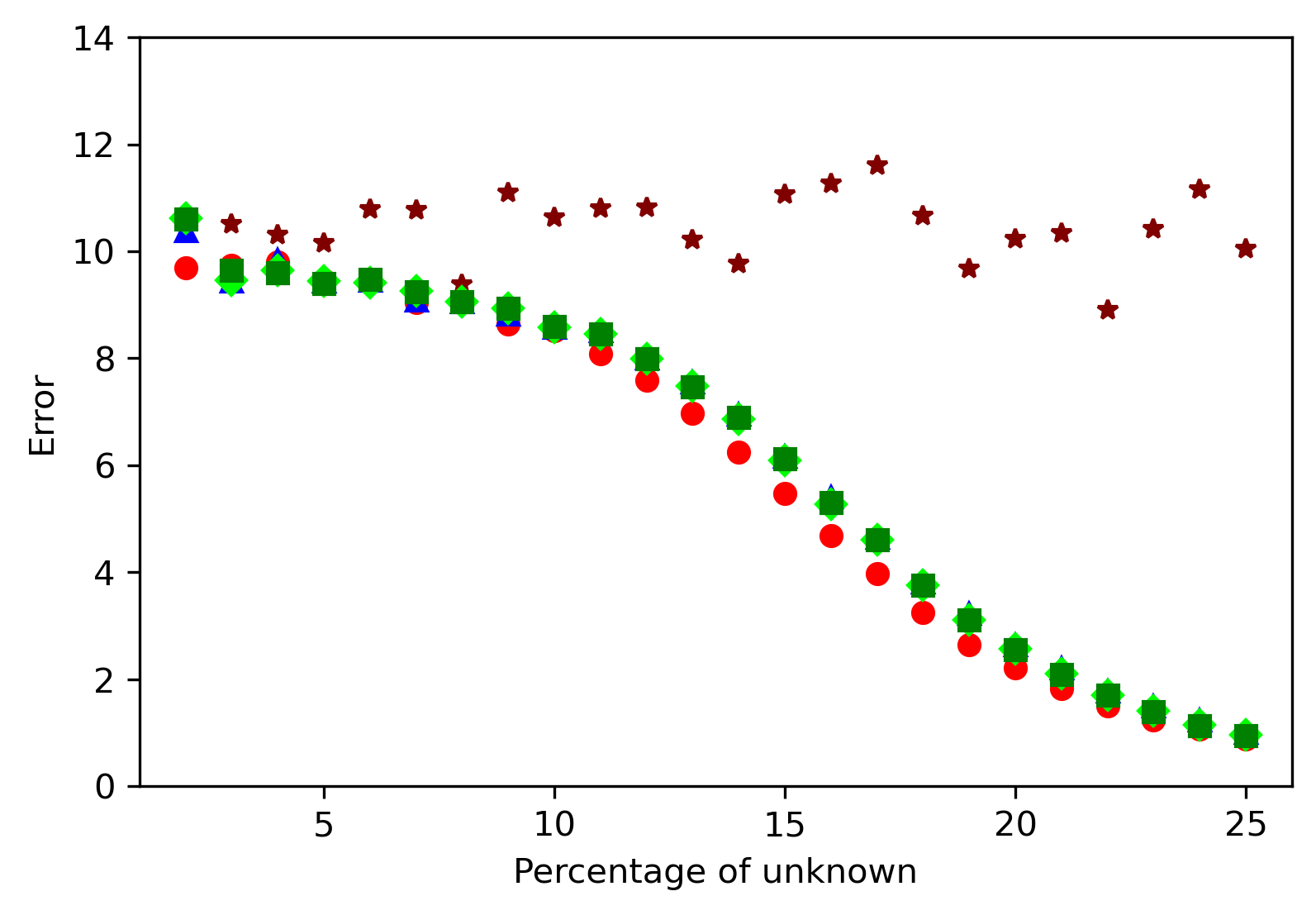}}
\caption{Performance of proposed policies when the threshold is selected to have less than 1\% early detection on validation test with 2\% unknown. Compare with Fig~\protect\ref{fig_performance_over_unknown} in main paper.}
\label{fig_performance_over_unknown_1early}
\end{figure}

\begin{figure}[htb!]
\centering
\subfloat[]{\includegraphics[width=2.5in]{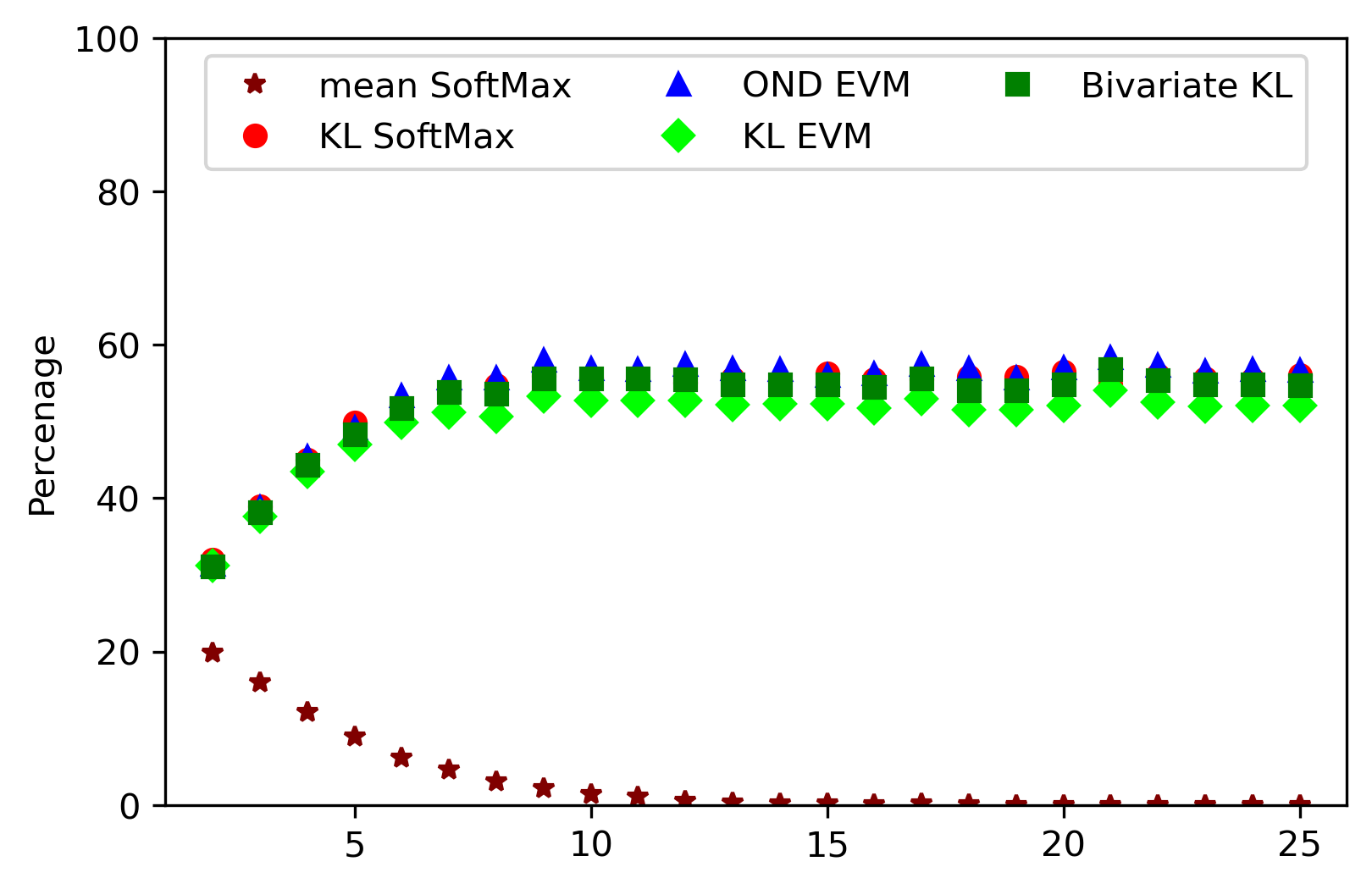}}\\
\subfloat[]{\includegraphics[width=2.5in]{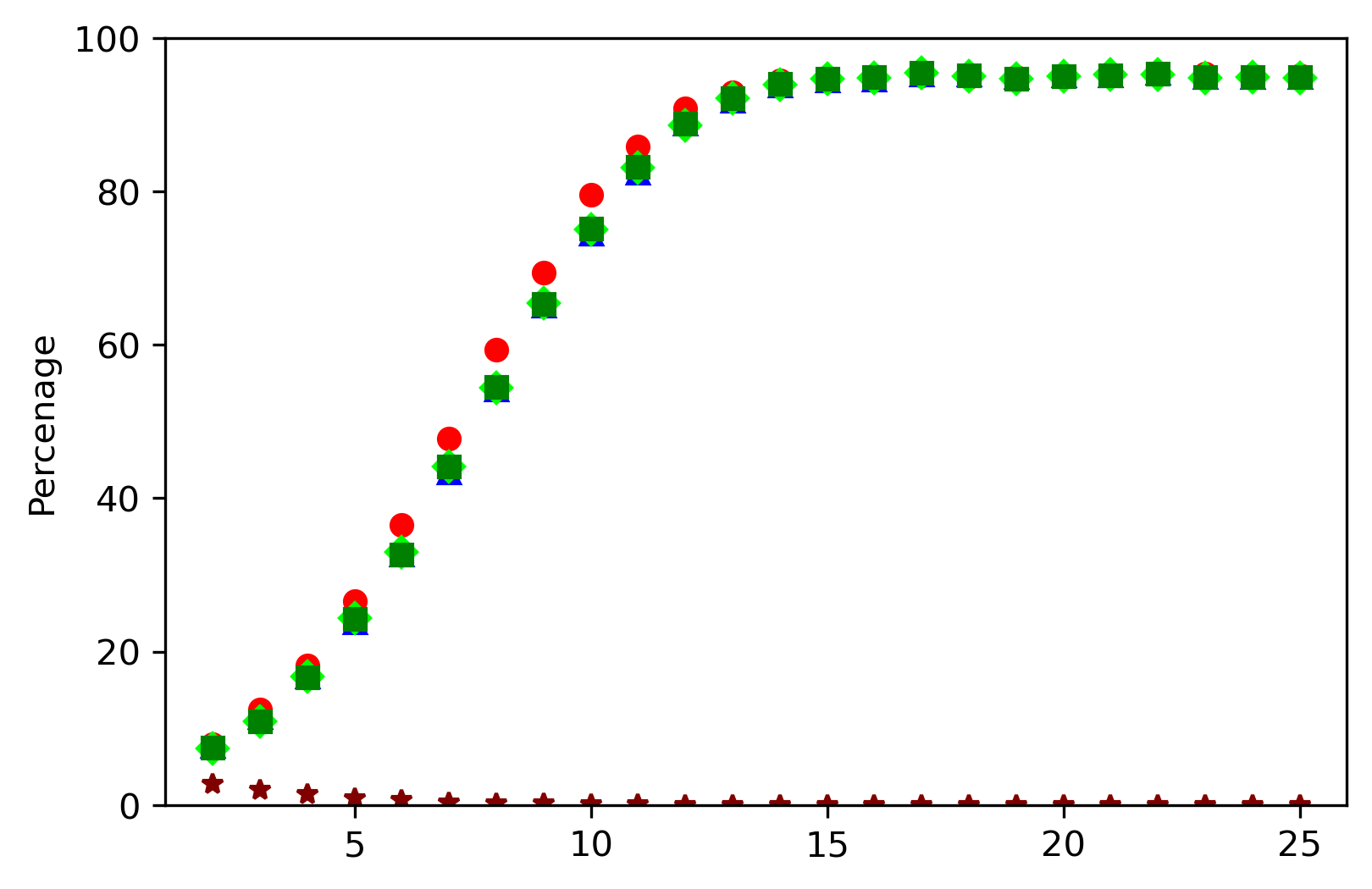}}\\
\subfloat[]{\includegraphics[width=2.5in]{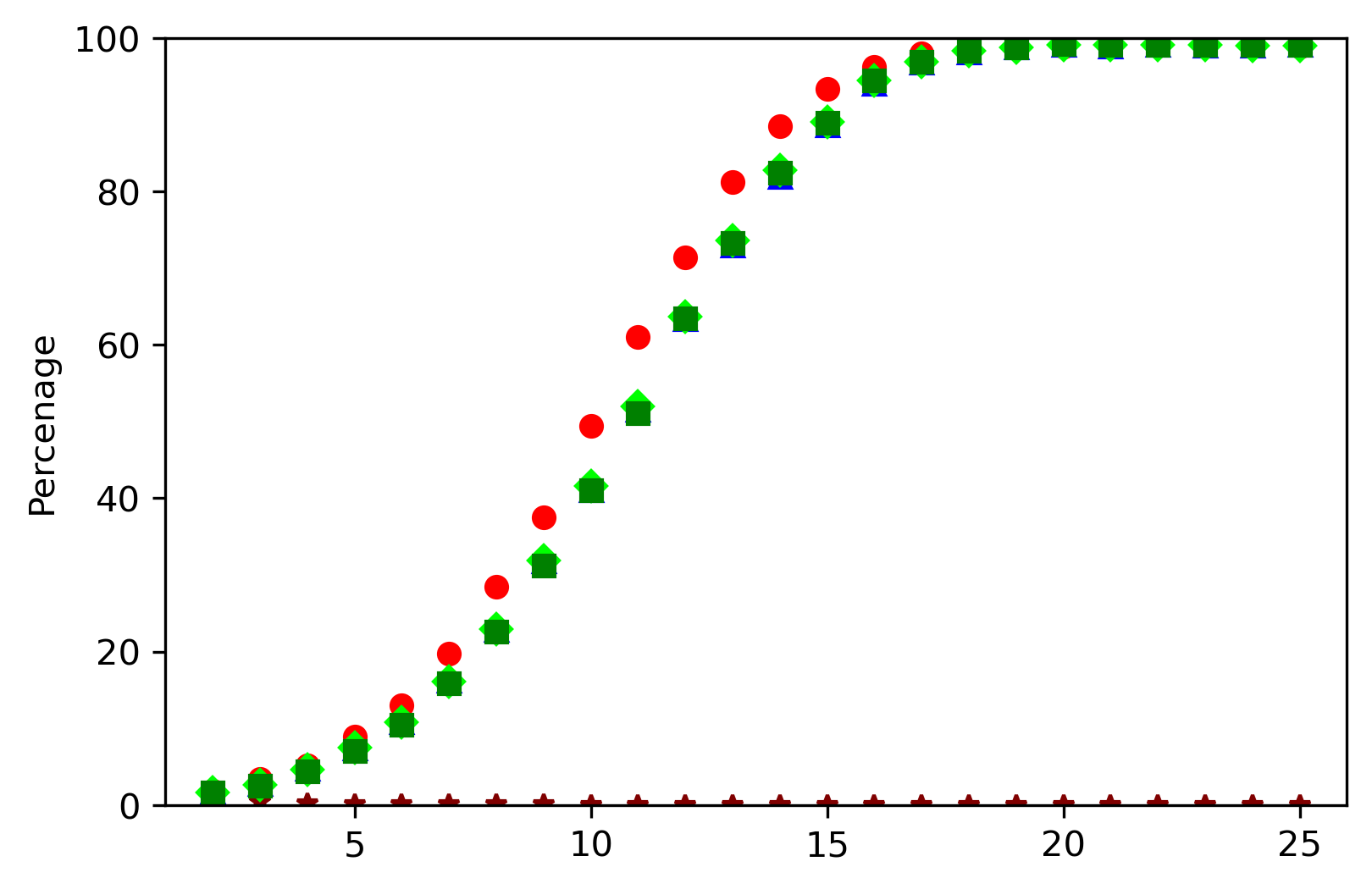}}\\
\subfloat[]{\includegraphics[width=2.5in]{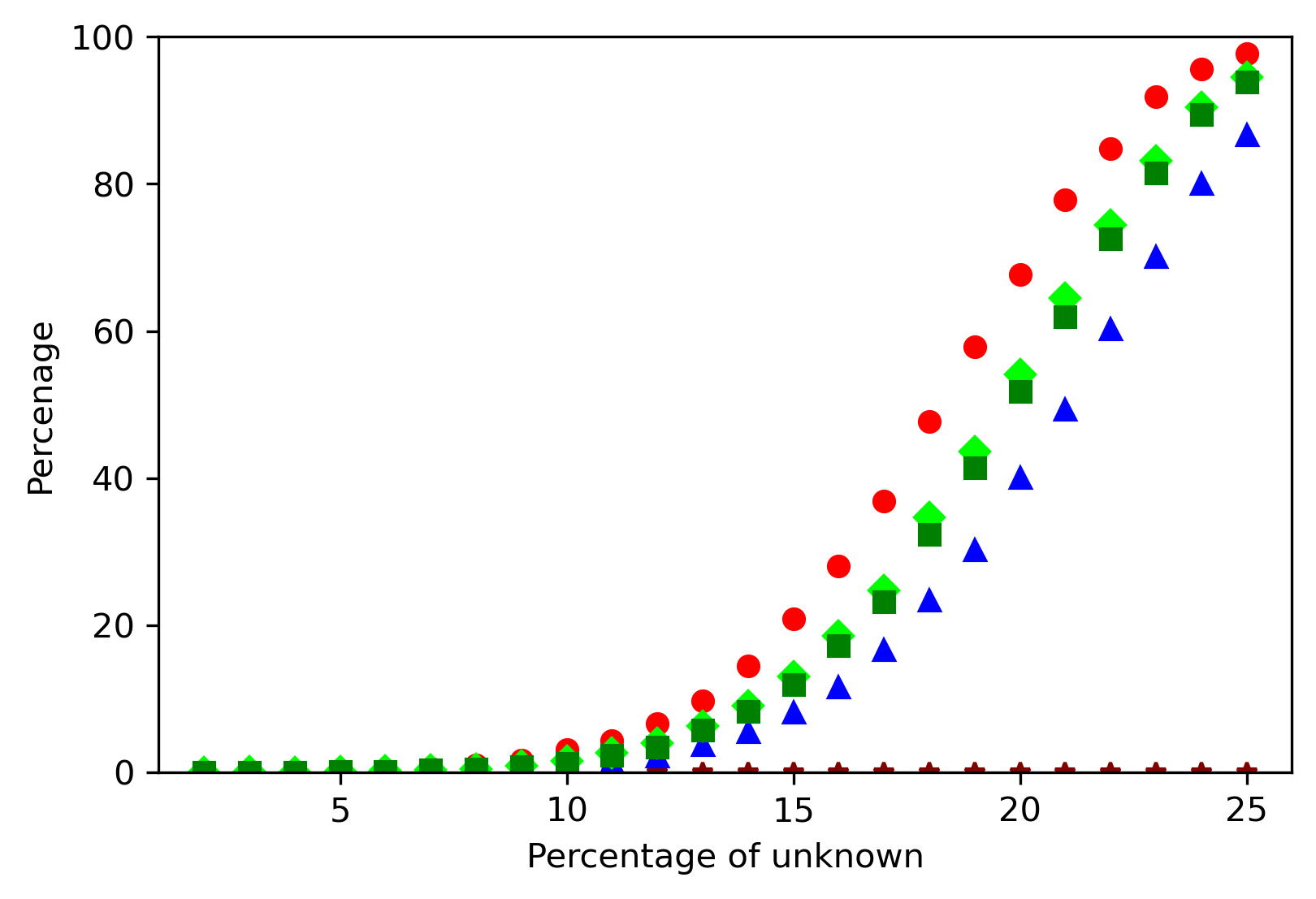}}
\caption{True detection percentage of proposed policies when the threshold is selected to (a) maximize true detection, (b) have 5\% early detection, (c) have 1\% early detection, (d) have not early detection on validation test with 2\% unknown.}
\label{fig_true_detection}
\end{figure}

\clearpage 

\begin{algorithm}[h]
\caption{Simplest (baseline) automatic reliability assessment of open-set image classifiers using mean of SoftMax}
\label{a:meanSoftMax}
\SetAlgoNoLine
\DontPrintSemicolon
\SetKwInput{KwData}{Config}
\SetKwInput{KwResult}{Initialize}
\KwIn{A batch of images, $\mu_{\textup{old}}$ state from past epoch (init 1.0)}
\KwData{$M$ tolerance to say image classifier is unreliable}
\KwOut{Reliability, $\mu$ state}
\;

\tcp{N: number of images in the batch}

x $\leftarrow$ normalize each images to range $[-1,+1]$

s $\leftarrow$ CNN(x) \tcp*{SoftMax, N x M}

p $\leftarrow$ $\max $ (s) \tcp*{Max over row, N x 1}

$\mu \leftarrow$ mean (p)

$\mu \leftarrow \min \{ \mu_{\textup{old}} , \mu \}$

\eIf{$\mu > M$}
{\Return (Reliable , $\mu$)}
{\Return (Unreliable , $\mu$)}
\end{algorithm}

\begin{algorithm}[h]
\caption{Information theory method for automatic reliability assessment of open-set image classifiers using Kullback–Leibler divergence of SoftMax}
\label{a:KLSoftMax}
\SetAlgoNoLine
\DontPrintSemicolon
\SetKwInput{KwData}{Config}
\SetKwInput{KwResult}{Initialize}
\KwIn{A batch of images, $\textup{D}_{\textup{old}}$ state from past epoch (init 0.0)}
\KwData{(m,s) mean and standard deviation of SoftMax of training data set, $\kappa$ tolerance to say image classifier is unreliable}
\KwOut{Reliability, D state}
\;

\tcp{N: number of images in the batch}

x $\leftarrow$ normalize each images to range $[-1,+1]$

s $\leftarrow$ CNN(x) \tcp*{SoftMax, N x M}

p $\leftarrow$ $\max $ (s) \tcp*{Max over row, N x 1}

$\mu \leftarrow$ mean (p)

$\sigma \leftarrow$ std (p)

D $\leftarrow$ KL($\mu$, $\sigma$, m , s) \tcp*{Equation \ref{eq_KL1}}

D $\leftarrow \max \{ \textup{D}_{\textup{old}}$ , D \}

\eIf{$\textup{D} < \kappa$}
{\Return (Reliable , D)}
{\Return (Unreliable , D)}
\end{algorithm}

\begin{algorithm}[h]
\caption{Proposed OND automatic reliability assessment using EVM open-set image classifier}
\label{a:ONDEVM}
\SetAlgoNoLine
\DontPrintSemicolon
\SetKwInput{KwData}{Config}
\SetKwInput{KwResult}{Initialize}
\KwIn{A batch of images, $\varepsilon_{\textup{old}}$ state from past epoch (init 0)}
\KwData{$\Delta$ lower bound limit of probability of EVM for image to be considered as known classes, $\hat{\rho}$ estimation of OOD class ratio, $\Xi$ tolerance to say image classifier is unreliable}
\KwOut{Reliability, $\varepsilon$ state}
\;

\tcp{N: number of images in the batch}
\tcp{M: feature size of CNN}
\tcp{L: number of known classes}

x $\leftarrow$ normalize each images to range $[-1,+1]$

f $\leftarrow$ CNN(x) \tcp*{Deep features, N x M}

P $\leftarrow$ EVM(f) \tcp*{Equation \ref{eq_EVM_p}, N x L}

p $\leftarrow$ $\max $ (P) \tcp*{Max over row, N x 1}

$\upsilon \leftarrow$ 1 - $\Delta$ - p \tcp*{N x 1}

$\nu \leftarrow \max \{ 0 , \upsilon \}$ \tcp*{element wise maximum}

$\mu \leftarrow$ mean ($\nu$)

$\zeta \leftarrow \mu - \hat{\rho} (1 - \Delta)$

$\eta \leftarrow \max \{ 0 , \zeta \}$

$\varepsilon \leftarrow \max \{ \varepsilon_{\textup{old}} , \eta \}$

\eIf{$\varepsilon < \Xi$}
{\Return (Reliable , $\varepsilon$)}
{\Return (Unreliable , $\varepsilon$)}
\end{algorithm}

\begin{algorithm}[h]
\caption{Proposed automatic reliability assessment of open-set image classifiers using Kullback–Leibler divergence of EVM}
\label{a:KLEVM}
\SetAlgoNoLine
\DontPrintSemicolon
\SetKwInput{KwData}{Config}
\SetKwInput{KwResult}{Initialize}
\KwIn{A batch of images, $\textup{D}_{\textup{old}}$ state from past epoch (init 0.0)}
\KwData{(m,s) mean and standard deviation of maximum class probability of EVM on training data set, $\kappa$ tolerance to say image classifier is unreliable}
\KwOut{Reliability, D state}
\;

\tcp{N: number of images in the batch}
\tcp{M: feature size of CNN}
\tcp{L: number of known classes}

x $\leftarrow$ normalize each images to range $[-1,+1]$

f $\leftarrow$ CNN(x) \tcp*{Deep features, N x M}

P $\leftarrow$ EVM(f) \tcp*{Equation \ref{eq_EVM_p}, N x L}

p $\leftarrow$ $\max $ (P) \tcp*{Max over row, N x 1}

$\mu \leftarrow$ mean (p)

$\sigma \leftarrow$ std (p)

D $\leftarrow$ KL($\mu$, $\sigma$, m , s) \tcp*{Equation \ref{eq_KL1}}

D $\leftarrow \max \{ \textup{D}_{\textup{old}}$ , D \}

\eIf{$\textup{D} < \kappa$}
{\Return (Reliable , D)}
{\Return (Unreliable , D)}
\end{algorithm}

\begin{algorithm}[tbh]
\caption{Proposed automatic reliability assessment of open-set image classifiers using bivariate Kullback–Leibler divergence of SoftMax and EVM}
\label{a:biKL}
\SetAlgoNoLine
\DontPrintSemicolon
\SetKwInput{KwData}{Config}
\SetKwInput{KwResult}{Initialize}
\KwIn{A batch of images, $\textup{D}_{\textup{old}}$ state from past epoch (init 0.0)}
\KwData{($m_1,m_2,s_1,s_2$) mean and standard deviation of maximum SoftMax and maximum class probability of EVM on training data set, $\kappa$ tolerance to say image classifier is unreliable}
\KwOut{Reliability, D state}
\;

\tcp{N: number of images in the batch}
\tcp{M: feature size of CNN}
\tcp{L: number of known classes}

x $\leftarrow$ normalize each images to range $[-1,+1]$

f, s $\leftarrow$ CNN(x) \tcp*{Deep features and SoftMax}

P $\leftarrow$ EVM(f) \tcp*{Equation \ref{eq_EVM_p}, N x L}

$\textup{p}_1 \leftarrow$ $\max $ (s) \tcp*{Max over row, N x 1}

$\textup{p}_2 \leftarrow$ $\max $ (P) \tcp*{Max over row, N x 1}

$\mu_1 \leftarrow$ mean ($\textup{p}_1$)

$\mu_2 \leftarrow$ mean ($\textup{p}_2$)

$\sigma_1 \leftarrow$ std ($\textup{p}_1$)

$\sigma_2 \leftarrow$ std ($\textup{p}_2$)

$\rho \leftarrow$ correlation ($\textup{p}_1 , \textup{p}_2$) 

\tcp{KL from equation \ref{eq_KL2_full} }
D $\leftarrow$ KL($\mu_1, \mu_2, \sigma_1, \sigma_2, \rho, m_1, m_2, s_1, s_2, r$)

D $\leftarrow \max \{ \textup{D}_{\textup{old}}$ , D \}

\eIf{$\textup{D} < \kappa$}
{\Return (Reliable , D)}
{\Return (Unreliable , D)}
\end{algorithm}

\end{document}